\documentclass[10pt,twocolumn,letterpaper]{article}
\usepackage{array} 
\usepackage[flushleft]{threeparttable}   
\usepackage{longtable}

\newif\ifhidepagenum
\hidepagenumfalse
\newif\ifusehyperref
\usehyperreftrue
\usepackage{titlesec}
\titlespacing*{\paragraph}{0pt}{.5ex plus .5ex minus .5ex}{1em}
\usepackage{iccv}
\usepackage{times}
\usepackage{epsfig}
\usepackage{graphicx}
\usepackage{amsmath}
\usepackage{amssymb}
\usepackage{color}
\usepackage{colortbl}
\usepackage{booktabs}
\usepackage[format=plain,labelformat=simple,labelsep=period,font=small,compatibility=false]{caption}
\usepackage{multirow}
\usepackage{makecell}
\usepackage{comment}
\usepackage{pifont}
\usepackage{enumitem}
\usepackage{overpic}
\usepackage[accsupp]{axessibility}
\usepackage{tabularx,stackengine,collcell}
\let\endminwd\relax
\newcolumntype{L}[1]{>{\collectcell\xminwd l{#1}}l<{\endminwd\endcollectcell}}
\newcolumntype{C}[1]{>{\collectcell\xminwd c{#1}}c<{\endminwd\endcollectcell}}
\newcolumntype{R}[1]{>{\collectcell\xminwd r{#1}}r<{\endminwd\endcollectcell}}
\def\minwd#1#2#3\endminwd{\stackengine{0pt}{#3}{\rule{#2}{0pt}}{O}{#1}{F}{F}{L}}
\newcommand\xminwd[1]{\minwd#1}

\usepackage{pgfplots}
\pgfplotsset{compat=1.9}

\ifusehyperref
\usepackage[pagebackref,breaklinks,colorlinks]{hyperref}
\else
\usepackage[pagebackref,breaklinks,colorlinks,bookmark=false]{hyperref}
\fi
\hypersetup{
	colorlinks,
	linkcolor={blue!80!black},
	citecolor={blue!80!black},
	urlcolor={blue!80!black},
    pdftitle={Annolid: Annotate, Segment, and Track Anything You Need},
}

\iccvfinalcopy 

\ifhidepagenum
    \ificcvfinal\fi
\fi

\makeatletter
\def\@fnsymbol#1{\ensuremath{\ifcase#1\or \dagger\or \ddagger\or
		\mathsection\or \mathparagraph\or \|\or **\or \dagger\dagger
		\or \ddagger\ddagger \else\@ctrerr\fi}}
\makeatother

\newcommand{\beginsupplement}{
	\setcounter{table}{0}
	\renewcommand{\thetable}{S\arabic{table}}%
	\setcounter{figure}{0}
	\renewcommand{\thefigure}{S\arabic{figure}}%
	\setcounter{equation}{0}
	\renewcommand{\theequation}{S\arabic{equation}}
}

\begin{document}

\title{Annolid: Annotate, Segment, and Track Anything You Need } 

\author{Chen Yang\textsuperscript{1} \hspace{1em} Thomas A. Cleland\textsuperscript{1} \hspace{1em}\\
\textsuperscript{1}Dept. of Psychology, Cornell University, Ithaca, NY 14853 \hspace{2em}\\
{\tt\small\{cy384,tac29\}@cornell.edu}
\vspace{-1ex}
}

\twocolumn[{%
\renewcommand\twocolumn[1][]{#1}%
\ifhidepagenum
    \pagenumbering{gobble} 
\fi
\maketitle
\centering
\captionsetup{type=figure}
\centering
\begin{tabular}{c@{\hspace{2pt}}c@{\hspace{2pt}}c}
\includegraphics[width=0.32\linewidth]{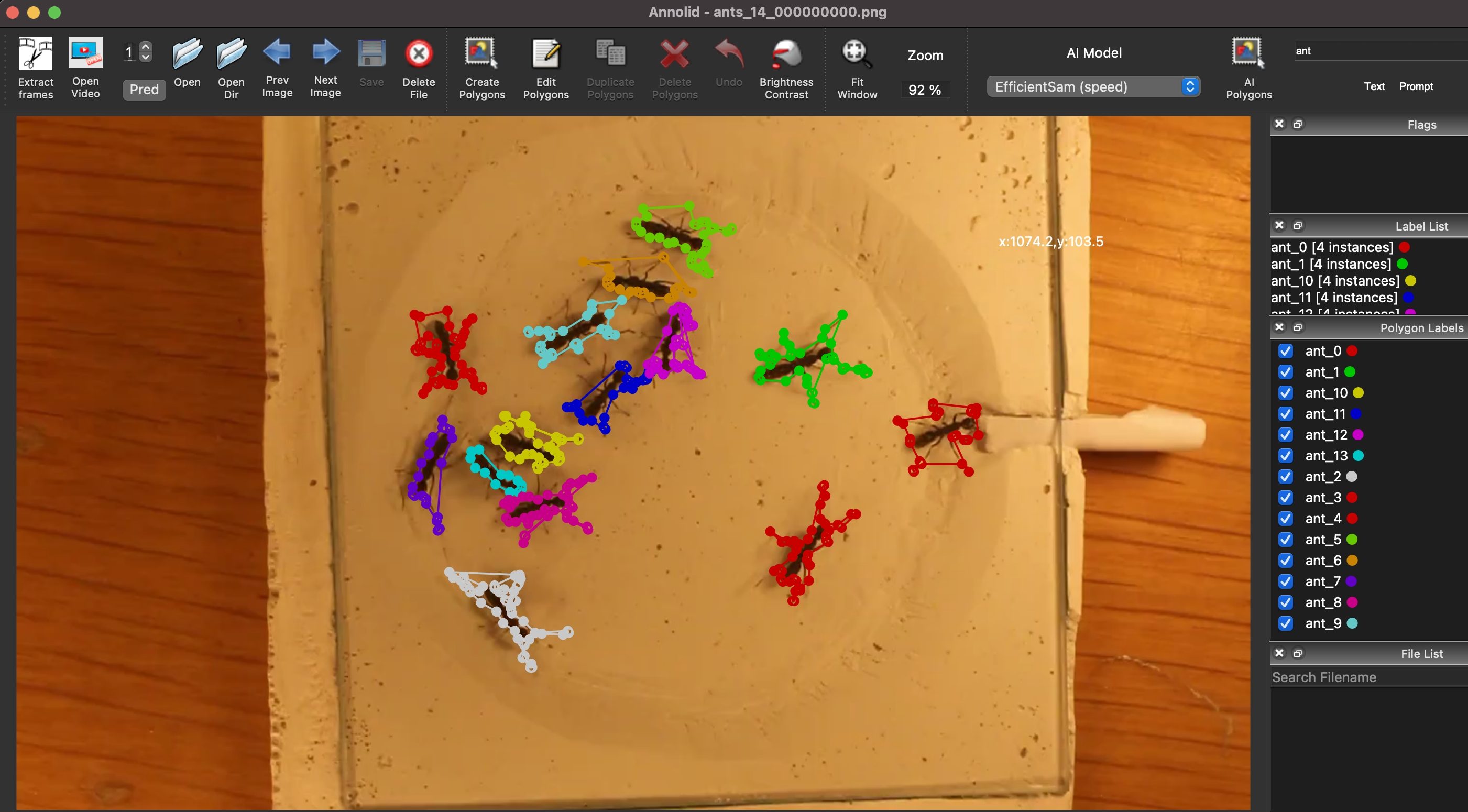} &
\includegraphics[width=0.32\linewidth]{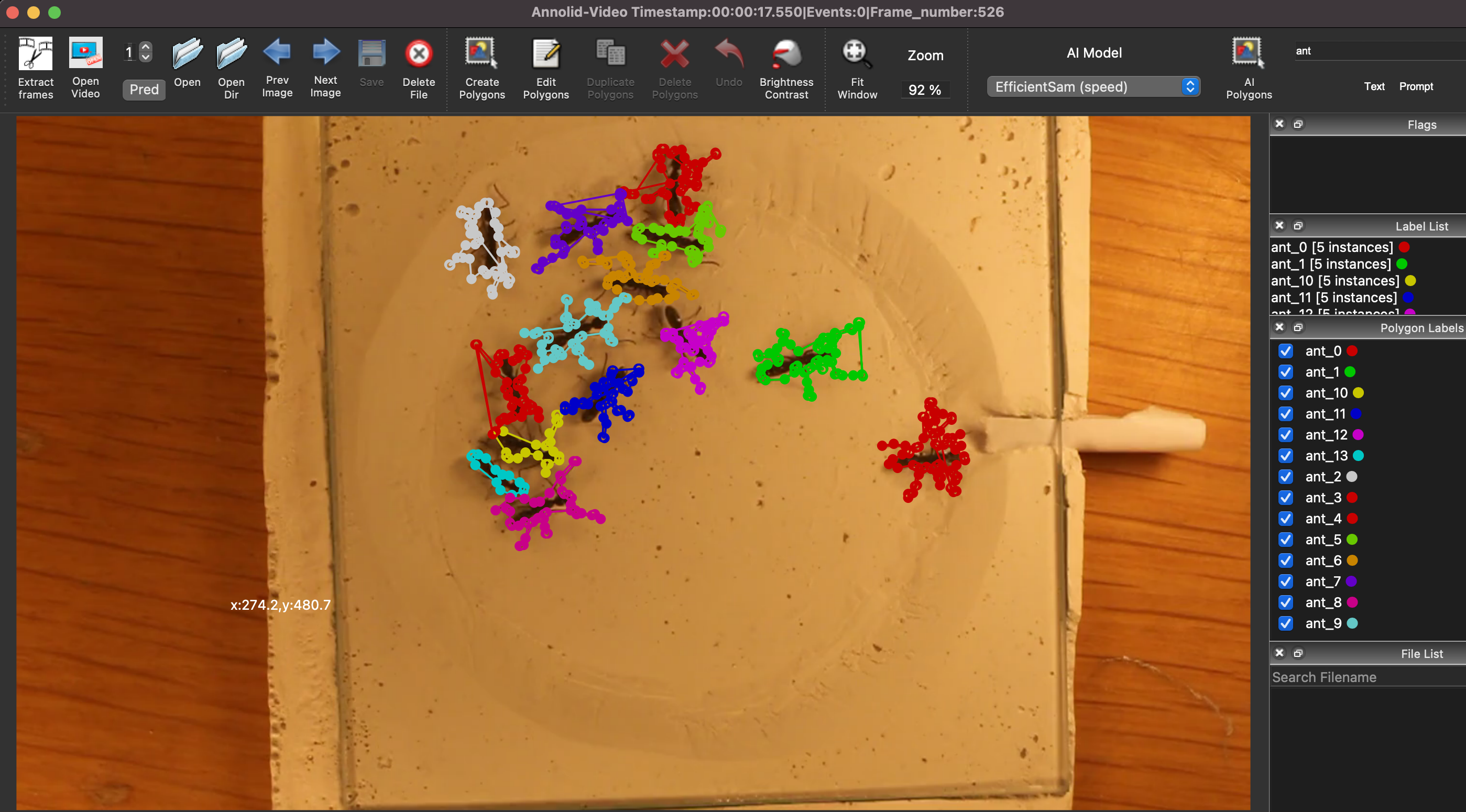} &
\includegraphics[width=0.32\linewidth]{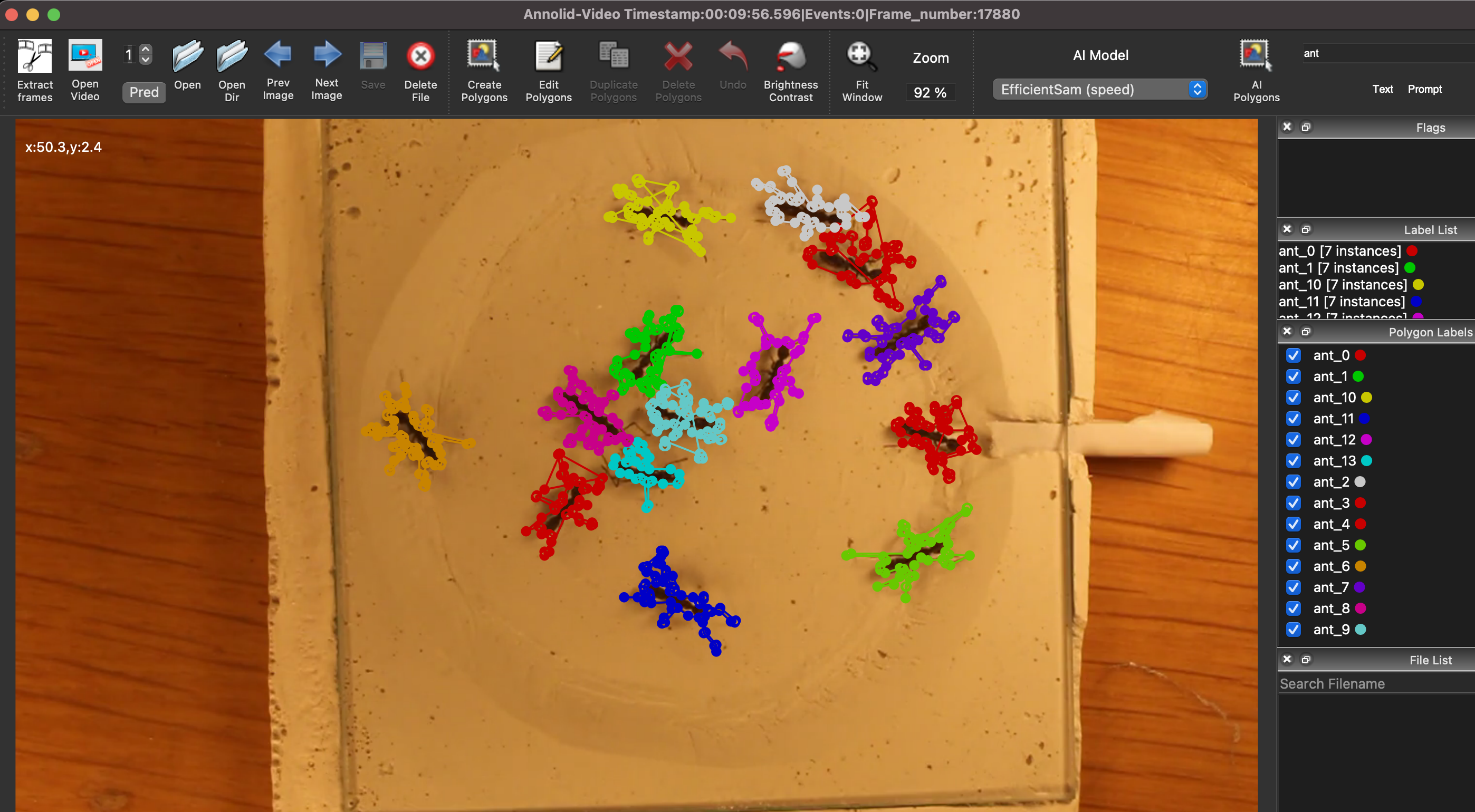} \\
\includegraphics[width=0.32\linewidth]{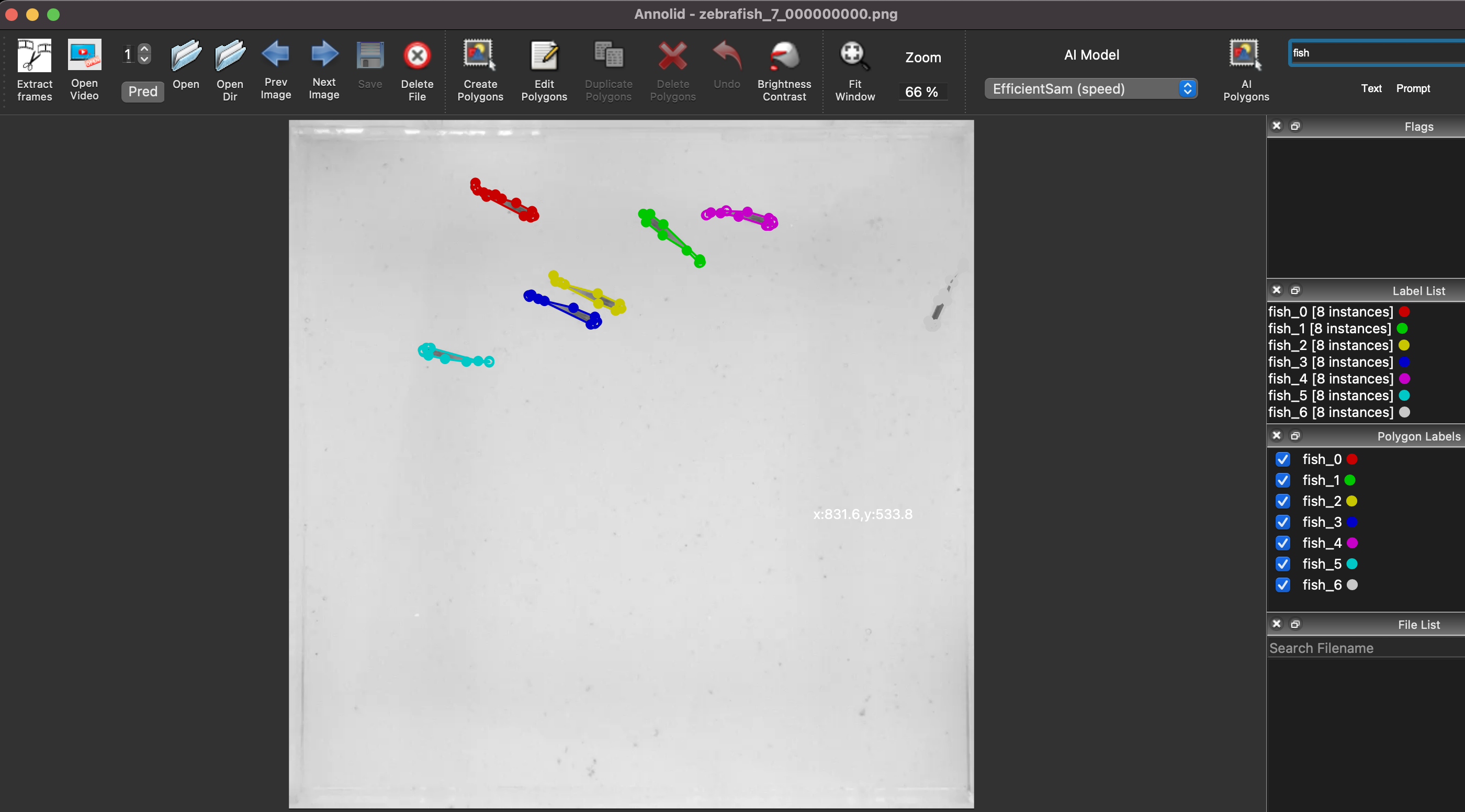} &
\includegraphics[width=0.32\linewidth]{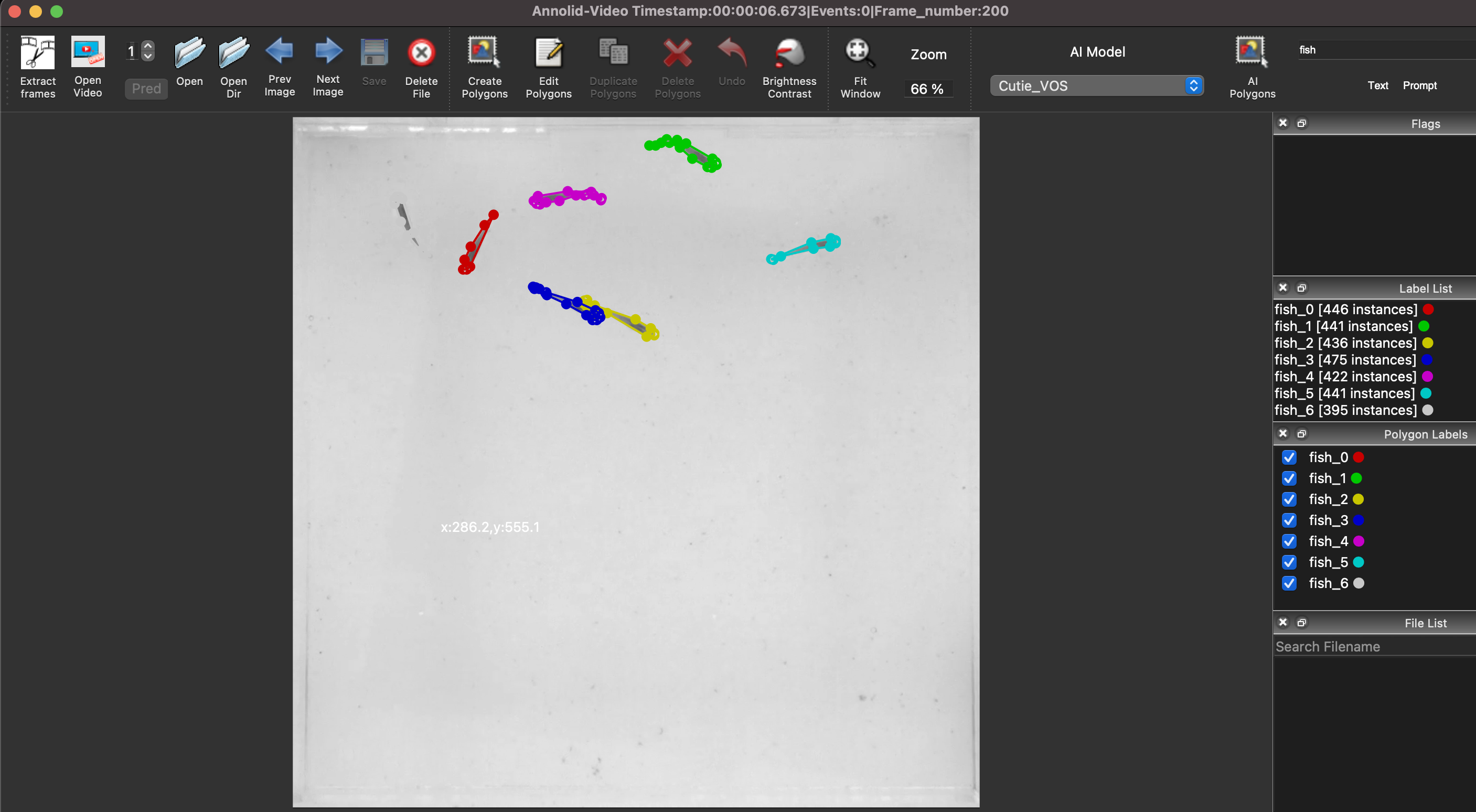} &
\includegraphics[width=0.32\linewidth]{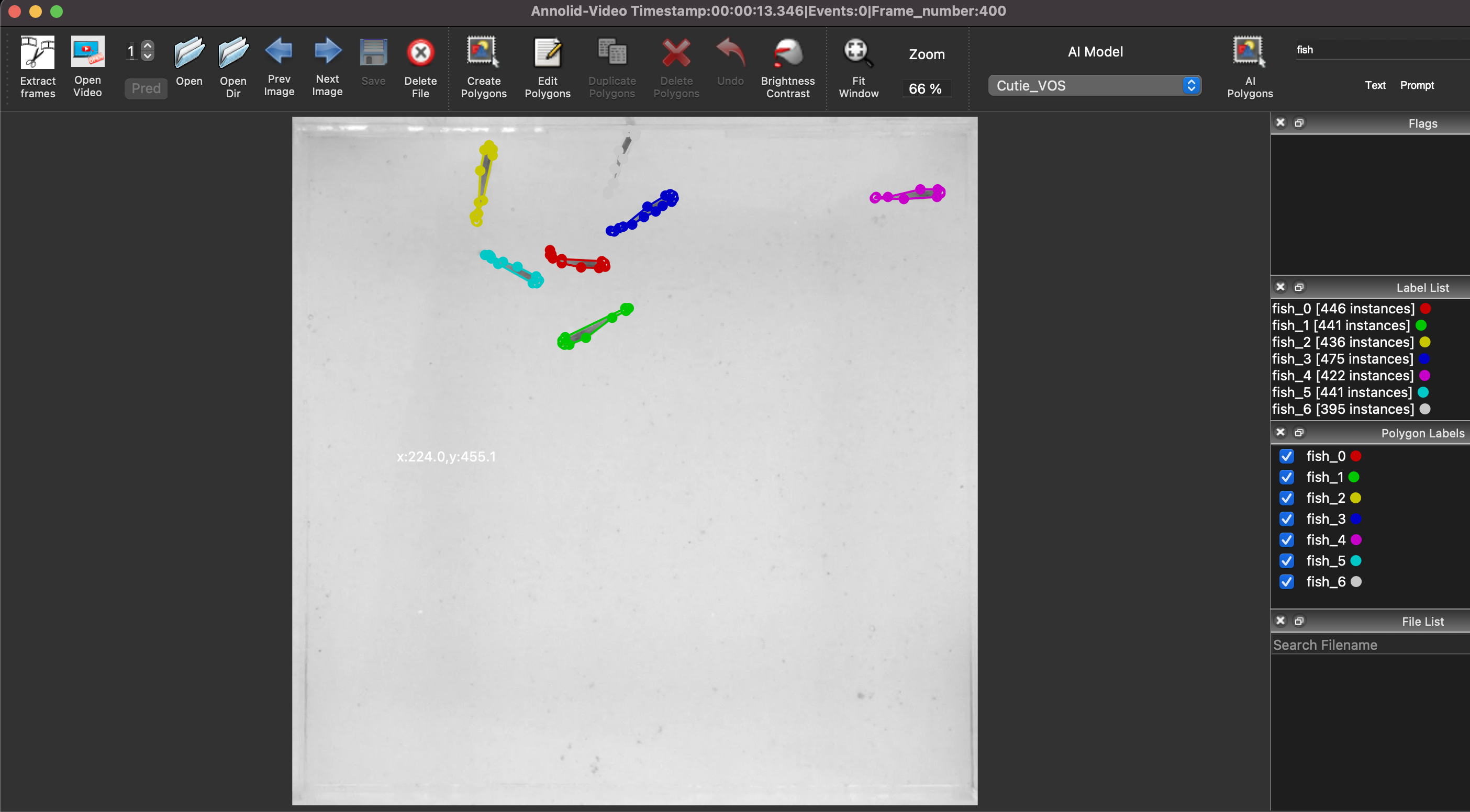} 
\end{tabular}
\begin{tabular}{c@{\hspace{2pt}}c@{\hspace{2pt}}c@{\hspace{2pt}}c@{\hspace{2pt}}c}
\includegraphics[width=0.19\linewidth]{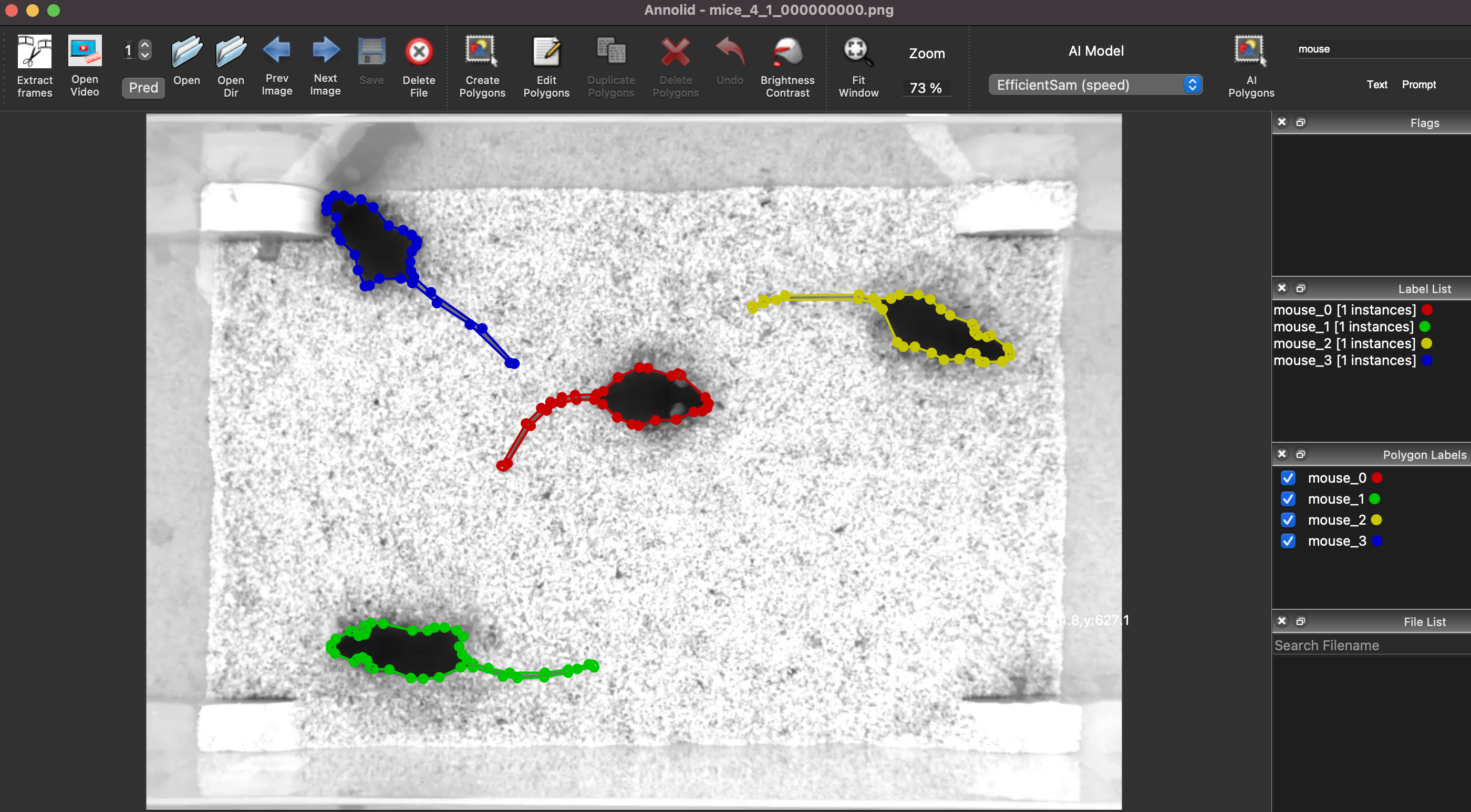} &
\includegraphics[width=0.19\linewidth]{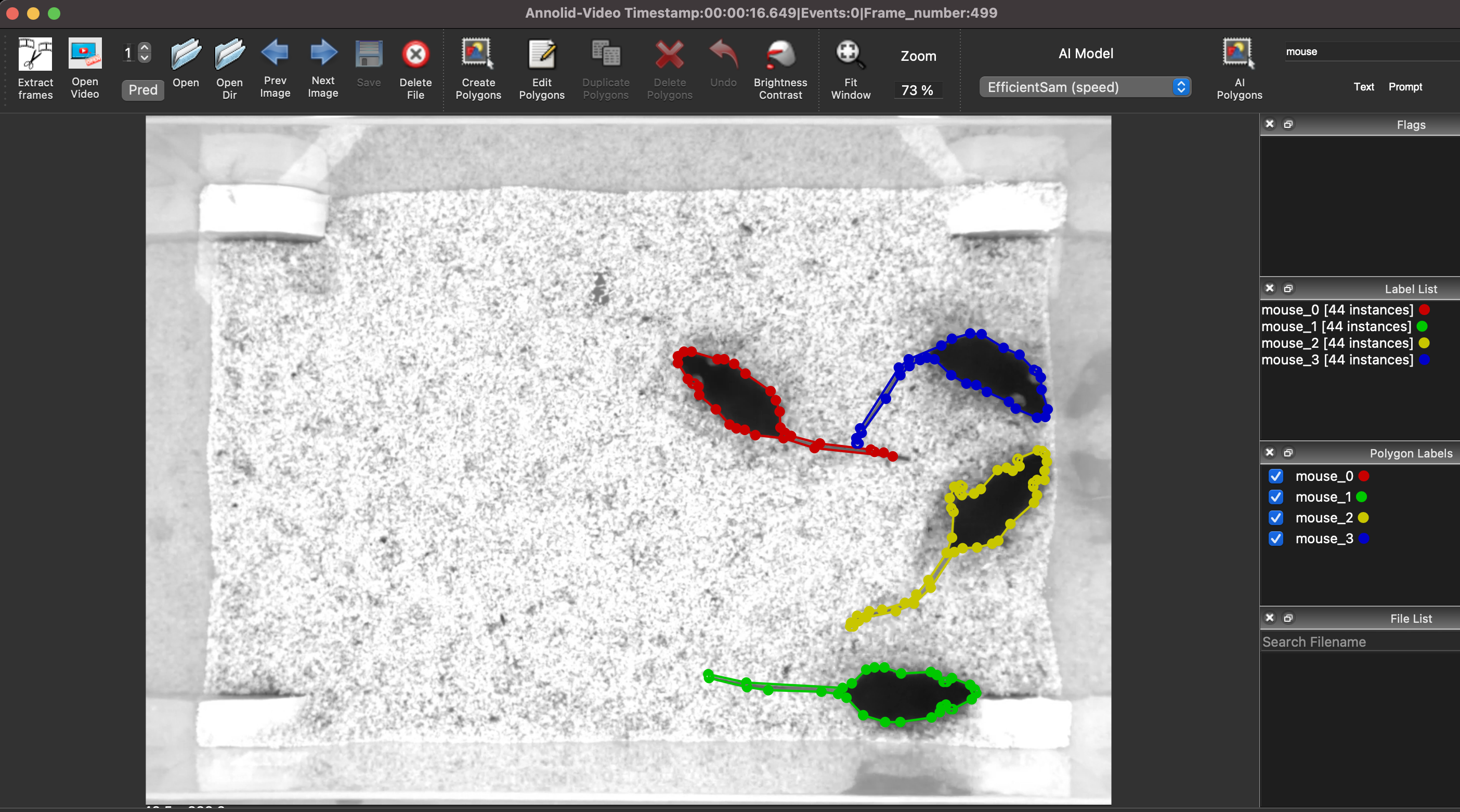}  &
\includegraphics[width=0.19\linewidth]{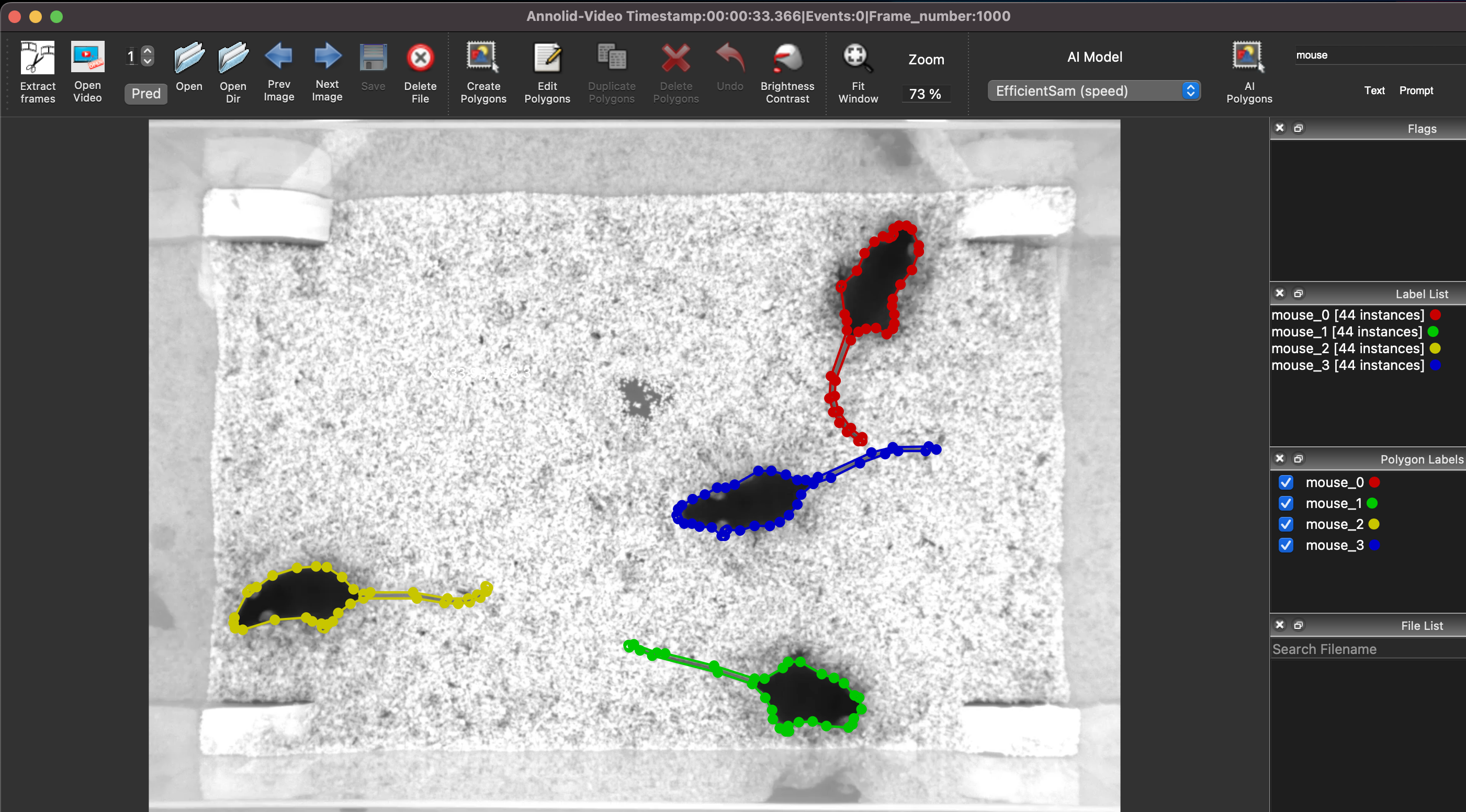} &
\includegraphics[width=0.19\linewidth]{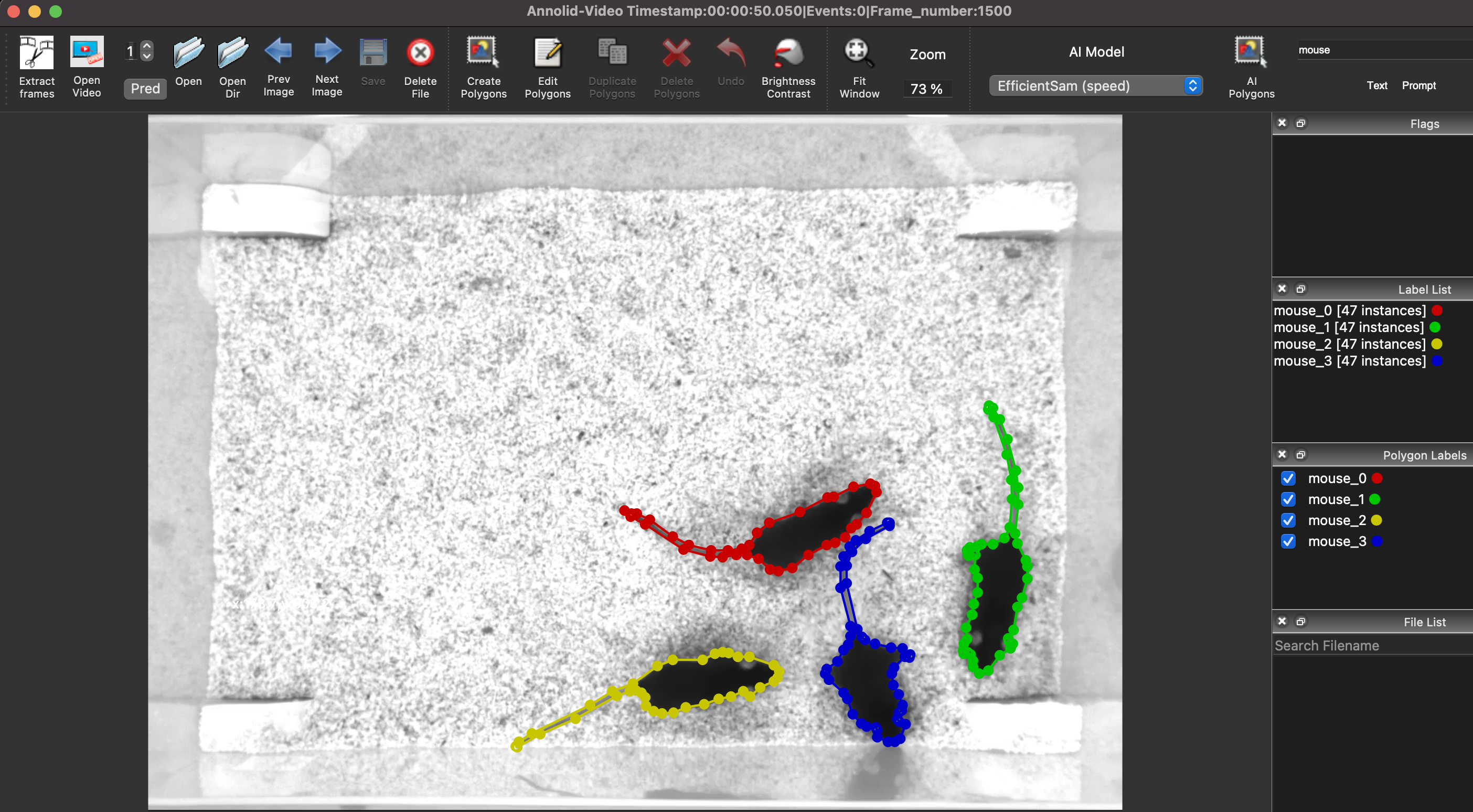}  &
\includegraphics[width=0.19\linewidth]{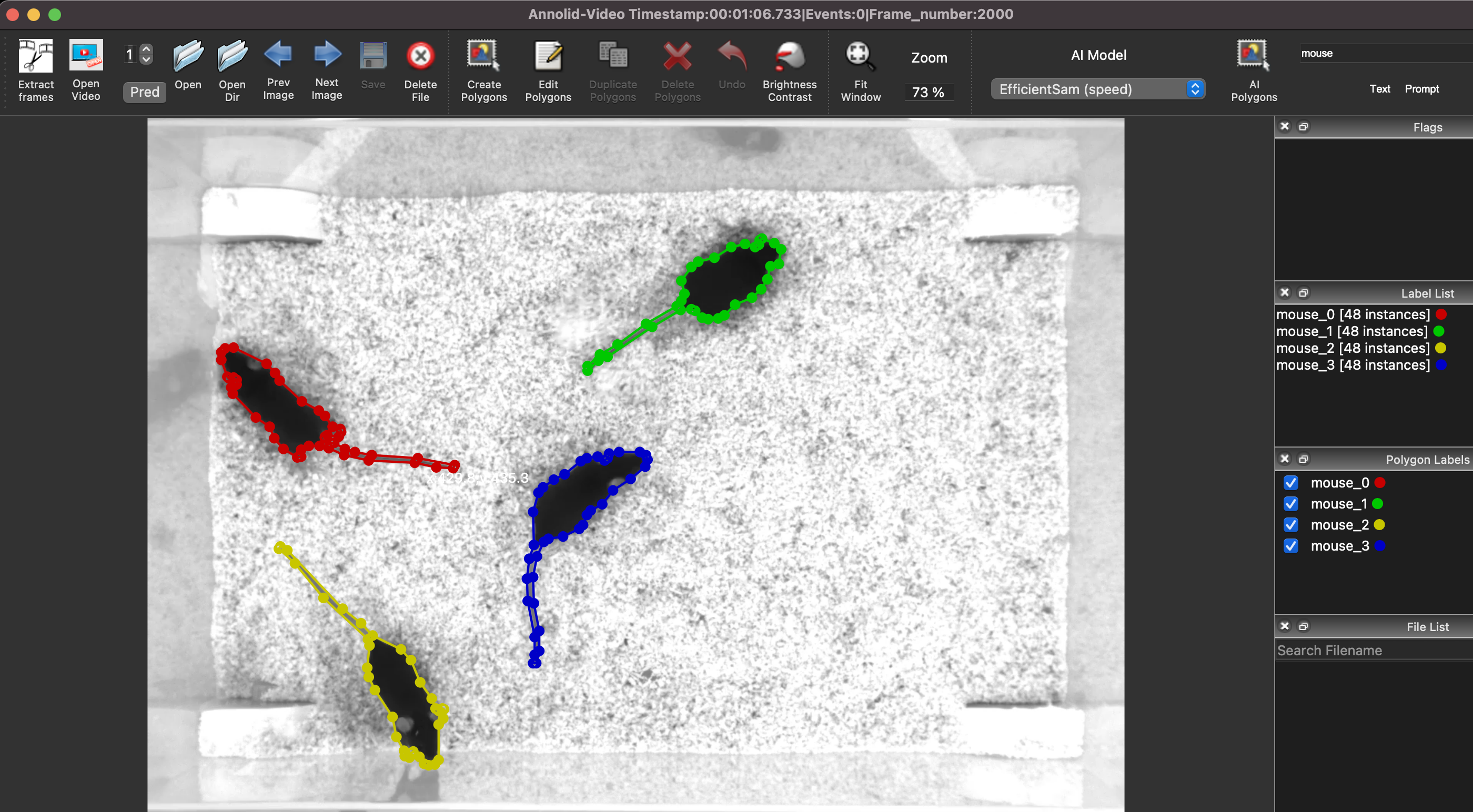} 

\end{tabular}
\captionof{figure}{
    \small 
    Examples of multiple markerless animal tracking results in Annolid \cite{yang2023automated}. Annolid now utilizes the Grounding-DINO  \cite{liu2023grounding} and Segment Anything \cite{kirillov2023segany, sam_hq, mobile_sam} models to automatically segment and label all instances of a named class in an initial frame, and then leverages the Cutie \cite{cheng2023putting} open-world video object segmentation (VOS) model to track multiple animals throughout video recordings based on that single labeled frame (zero-shot learning).  
    \textbf{Top:} Based on the end user entering the text \textit{``ant''} in the search field at the upper right, Annolid automatically segments all instances matching that label (i.e., ants) in the initial frame (\textit{left panel}), and then tracks the labeled animals across frames throughout the video (\textit{middle and right panels}).
    \textbf{Middle:} As in the top panel, except that seven zebrafish are tracked based on a single frame of autolabeled instances (text prompt \textit{``fish''}).
    \textbf{Bottom:} As in the top panel, except that four mice are tracked based on a single frame of autolabeled instances (text prompt \textit{``mouse''}).  Annolid successfully tracked the mice and ants throughout each ten-minute video using only the polygons automatically generated in the first frame; zebrafish also were successfully tracked after incorporating human-in-the-loop corrections. Images are derived from videos in the \textit{idTracker.ai} dataset~\cite{romero2019idtracker}. 
    \vspace{0.3 cm}
    }%
\label{fig:teaser} 
\vspace{0.5ex}   
}]

\begin{abstract}
    Annolid is a deep learning-based software package designed for the segmentation, labeling, and tracking of research targets within video files, focusing primarily on animal behavior analysis. Based on state-of-the-art instance segmentation methods, Annolid now harnesses the Cutie video object segmentation model to achieve resilient, markerless tracking of multiple animals from single annotated frames, even in environments in which they may be partially or entirely concealed by environmental features or by one another. Our integration of Segment Anything and Grounding-DINO strategies additionally enables the automatic masking and segmentation of recognizable animals and objects by text command, removing the need for manual annotation. Annolid's comprehensive approach to object segmentation flexibly accommodates a broad spectrum of behavior analysis applications, enabling the classification of diverse behavioral states such as freezing, digging, pup huddling, and social interactions in addition to the tracking of animals and their body parts.

\end{abstract}
\section{Introduction}

The field of animal behavior analysis is hugely diverse, requiring a broad panoply of strategies for identifying and scoring specific aspects of complex behavior exhibited by individual animals and groups.  While software-assisted methods for scoring behavior are in widespread use, some types of behavior analysis are much better served than others by existing tools and strategies.  We initially developed Annolid \cite{yang2023automated} to better address this great diversity of challenges.  Uniquely among deep learning-based behavior analysis packages, Annolid employs a strategy of instance segmentation, treating each instance as a distinct class.  For example, in multi-animal tracking applications, individual animals correspond to separate instances, and hence can be distinguished and tracked despite periods of occlusion or close interactions with other animals. Instances also can correspond to particular behavioral states, such as grooming, digging, or huddling in groups. In the original Annolid pipeline \cite{yang2023automated}, end users identify instances by drawing polygons (instance masks) on some number of video frames within the Annolid GUI, based on LabelMe~\cite{Wada_Labelme_Image_Polygonal}. These labeled polygons then are converted into a COCO format dataset for transfer learning purposes, and used to train an instance segmentation network, typically a Mask R-CNN~\cite{he2017mask} network implemented in Detectron2~\cite{wu2019detectron2}.  The trained model subsequently is used for inference on all frames within a video, thereby tracking animals (or specific body parts), identifying epochs of particular behavior states or interactions, and/or measuring motion between frames~\cite{yang2023automated}.  Annolid requires relatively few training frames for high performance, and supports an iterative human-in-the-loop strategy to help focus end users' labeling efforts on the more difficult generalizations in a given video.  However, the time and effort required to score frames remain the limiting factors in most behavior analysis pipelines, and become increasingly burdensome as the number of instances (e.g., animal group size) increases.  

In response to this challenge, we here introduce a fundamentally new strategy for object segmentation and tracking based on the integration into Annolid of three transformative machine learning tools.  First, the incorporation of Cutie~\cite{cheng2023putting}, a cutting-edge video object segmentation (VOS) model, enables Annolid to accurately predict and segment up to ~100 separate instances across the full duration of a video recording based on a single labeled frame. Briefly, Cutie's VOS strategy propagates the instance masks and identities defined by the first labeled "ground truth" frame by maintaining a multi-frame memory buffer that  integrates and utilizes both pixel- and object-level memory to predict instances across frames. Among other benefits, this tracking-centric method effectively eliminates "teleportation" arising from identity switches. 
 
Second, we incorporate Meta AI's Segment Anything Model (SAM)~\cite{kirillov2023segany,sam_hq}, which enables the automatic masking of visually discrete objects via zero-shot generalization -- i.e., without any need to manually specify objects (outline with polygons) or to train an Annolid model.  Third, to inform Annolid which objects in a video frame should be automatically segmented and labeled, we incorporate Grounding-DINO~\cite{liu2023grounding}, an open-set object detector that can identify arbitrary objects in a visual scene based upon text-based descriptors such as category names.  The combination of these latter two models enables end users to, for example, enter the category label "mouse" in the provided text field, after which Annolid will automatically segment and label all mice in the designated video frame as separate instances (Figure~\ref{fig:teaser}).  This initial labeled frame then can be utilized by Annolid's Cutie model to predict and segment those instances (i.e., track those mice) across all frames in the video.  Importantly, the distinct capabilities of the three new models also can be separately applied; for example, esoteric objects that are not recognized by name still can be manually annotated and labeled in a single frame, and then  predicted and segmented throughout the video file without the need to train multiple frames.  Similarly, complex, user-defined instances that do not follow natural visual segmentation boundaries \cite{fang2023annolid} still can be specified and segmented through the traditional annotation of frames followed by model training. 

In summary, Annolid now enables end users to specify animals and objects of interest in an initial video frame by name, and then will automatically identify and label all specified instances and propagate the resulting masks from the initial frame throughout the extent of the video file without the need to manually annotate any frames or train an Annolid model.  Annolid additionally provides methods by which any errors in this process can be easily identified and edited by end users via an iterative human-in-the-loop correction procedure. This consolidated strategy effectively transfers knowledge from large-scale and open-world datasets into the process of animal behavior analysis, greatly reducing end user effort while achieving state-of-the-art results on animal behavior datasets including Annolid's Multiple Animal Tracking \& Behavior (MATB) collection ~\cite{yang2023automated} and the video repository assembled by the \textit{idtracker.ai} project~\cite{romero2019idtracker}.

\section{Methods}

\subsection{Computational Environment}

Experiments were conducted on an Intel Core i9 workstation with an NVIDIA RTX3090 GPU. On this workstation, Annolid achieved near-real-time inference and playback speeds during the process of prediction at peak accuracy settings. Additional testing was conducted on a MacBook Air lacking a discrete GPU, which proved adequate for annotations using the GUI and for optimizing parameters such as \textit{mem\_every} and $T_{max}$ with short video samples. The pretrained Cutie model \texttt{cutie-base-mega.pth} (release v1.0) was obtained from the Cutie GitHub repository (\url{https://github.com/hkchengrex/Cutie}) \cite{cheng2023putting}. Throughout the study, default Cutie hyperparameters were utilized, with the exception of our exploration into optimizing tracking performance by varying the \textit{mem\_every} parameter. Pretrained SAM weights were downloaded from official GitHub releases (SAM \cite{kirillov2023segany}, SAM-HQ \cite{sam_hq}, and EdgeSAM \cite{zhou2023edgesam}); the analyses herein use the SAM-HQ model (\url{https://huggingface.co/lkeab/hq-sam/blob/main/sam_hq_vit_l.pth}). The Grounding DINO ONNX weights file (release 1.0.0) was downloaded from X-AnyLabeling (\url{    https://github.com/CVHub520/X-AnyLabeling/releases/download/v1.0.0/groundingdino_swinb_cogcoor_quant.onnx}) \cite{X-AnyLabeling}.

\subsection{Data Sources and Validation}

We selected videos for analysis from Annolid's Multiple Animal Tracking \& Behavior (MATB) collection ~\cite{yang2023automated}, assembled at ~\url{https://cplab.science/matb}, and from the \textit{idtracker.ai} dataset~\cite{romero2019idtracker}, available from a Google Drive-based data repository linked from \url{https://idtracker.ai} and identified below by their filenames in that repository. The videos in the MATB repository comprise an array of tracking challenges, focusing on natural complexities (such as object occlusion during interactions, unusually complex motion, and instances that vanish and reappear) and technical difficulties (such as camera motion, low contrast, visually noisy backgrounds, and events or reflections that impair the view of moving animals). These data also are used to test the automatic classification of specified behavioral states.  The videos in the \textit{idtracker.ai} dataset present an array of laboratory animal behavior scenarios featuring groups of animals (mice, ants, \textit{Drosophila} fruit flies, zebrafish) moving freely and interacting within various enclosures. 

We conducted tests on several video recordings, with durations of up to 10 minutes, on which the original Cutie model was not evaluated. To validate the accuracy of automatic tracking, we manually inspected the overlay of the generated masks on the animals using the Annolid viewer, reviewing each video frame by frame in its entirety.

\subsection{Operational Principles of Grounding DINO and SAM}

Grounding DINO \cite{liu2023grounding} is an open-set object detector that combines the Transformer-based detector DINO with grounded pre-training, enabling the detection of arbitrary objects based on human-provided text descriptors such as category names or descriptive expressions. The Segment Anything Model (SAM) \cite{kirillov2023segany} enables the automatic segmentation of images (i.e., the identification and masking of visual objects within scenes) based on an enormously rich pretrained model in which over 1 billion masks were created on 11 million images. SAM also is designed to be promptable with boxes and points; for example, Annolid's \textit{AI-Polygon} menu item uses SAM to specifically segment (draw a polygon around) individual visual objects that the user specifies with a single point (mouse click) in the Annolid GUI, a technique termed \textit{point prompting}.  This generative capacity greatly facilitates zero-shot transfer to new image distributions and tasks.  Annolid now combines the capabilities of these two models to enable the automatic segmentation of multiple animals in videos -- outlining each with an editable Annolid polygon -- based on a text prompt.  

\subsection{Automatic Object Detection and Segmentation in Annolid}

\begin{figure*}[t]
    \centering
    \includegraphics[width=\linewidth]{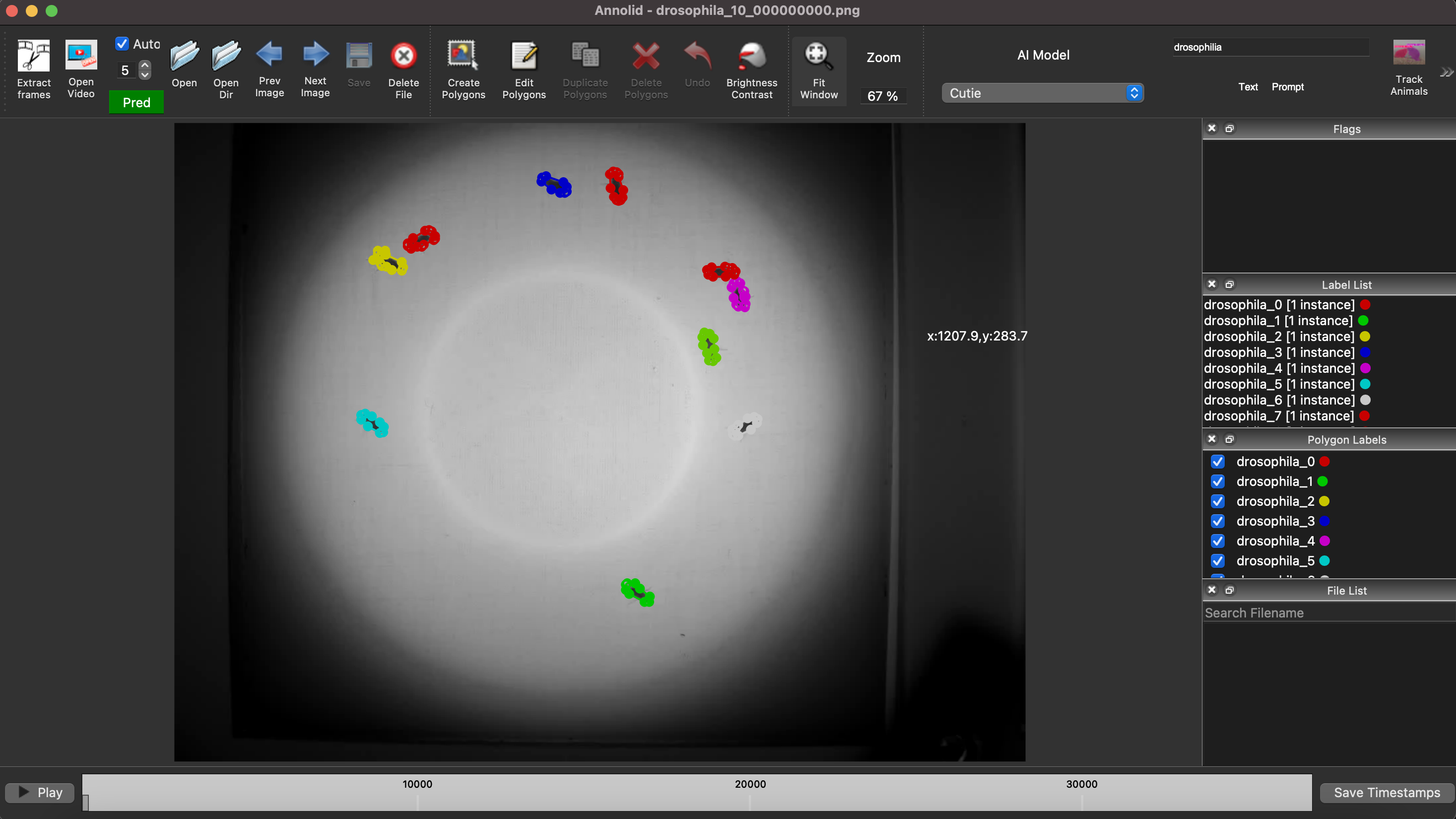}
    \caption{Illustration of the Annolid GUI, and elements of its labeling, prediction, and validation workflow. The top row features a set of GUI tools including an open video button and a spin box for setting the \textit{mem\_every} parameter before initiating the prediction process with the \textit{Pred} button. The text prompt box accepts words or phrases that define the automatic generation of polygons in the currently selected frame. Predicted polygons can be corrected manually, and labeling is saved in the LabelMe JSON file format.}
    \label{fig:annolid_gui}
\end{figure*}

Users generally will first downsample and compress the video file to improve computational efficiency.  An appropriate category label (e.g., "mouse") then is entered into the Annolid text prompt (upper right corner of the GUI; Figure~\ref{fig:annolid_gui}), after which Annolid automatically segments and labels instances in the initial video frame that are described by that category label, outlining each individual mouse with a polygon and assigning a unique class name based on the category label (e.g., "mouse\_1").  In cases where automatic segmentation misses some animals, or fails more broadly (as can be the case with unusual categories), users have the option to directly correct and/or create polygons using the Annolid GUI tools (e.g., via point prompting, described above, or by explicitly outlining instances).  Once all desired instances are correctly annotated and labeled in the initial frame, one proceeds to video object segmentation to propagate these instances across all frames of the video file.

\subsection{Operational Principles of Cutie Video Object Segmentation}
\label{section:cutieprinciples}

\begin{figure*}
    \centering
    \includegraphics[width=0.95\linewidth]{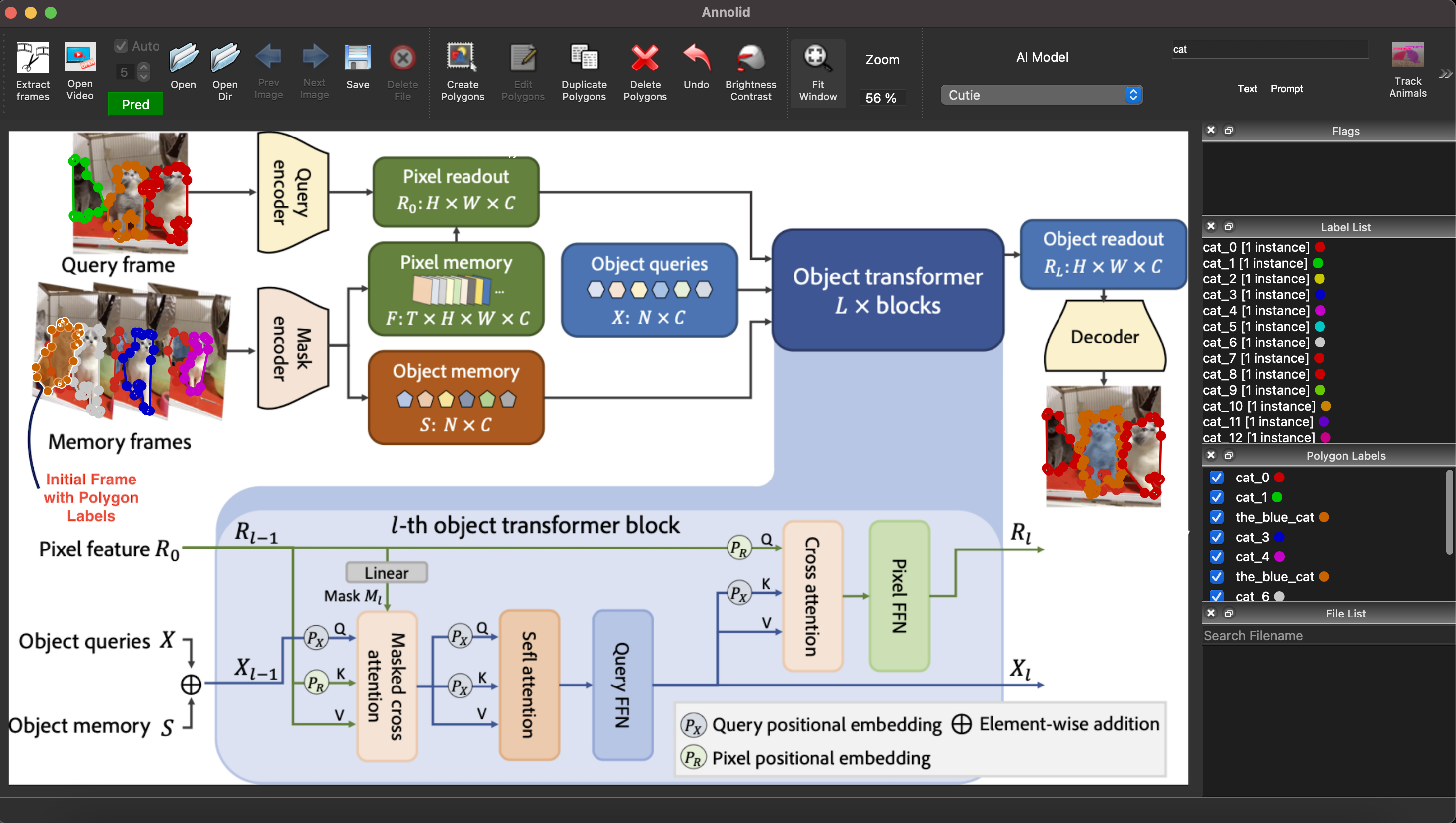}
    \caption{Overview of the Cutie architecture \cite{cheng2023putting} as integrated into Annolid. Labeled polygons are converted into masks from the currently selected frame and then stored in the FIFO memory buffer: specifically, pixel memory $F$ and object memory $S$, representing past segmented frames. Pixel memory is retrieved for the query frame as pixel readout $R_0$, which bidirectionally interacts with object queries $X$ and object memory $S$ in the object transformer. The object transformer comprises $L$ blocks that enrich the pixel features with object-level semantics and generate the final $R_L$ object readout for decoding into the output mask. Subsequently, the output mask is converted back to polygons for easy editing and visualization in the Annolid GUI.}
    \label{fig:overview}
\end{figure*}

The Cutie VOS model operates in a fully online manner, sequentially segmenting subsequent frames in a video file based on the segmentation and labeling of target objects in a single annotated frame. To do this, the Cutie architecture leverages end-to-end object-level information and bidirectional communication between pixel-level and object-level features. For example, Cutie enriches pixel features with object-level semantics to produce final object readouts for decoding into output masks. Key components in this process include pixel memory $F$, object memory $S$, object queries $X$, and an object transformer comprising $L$ transformer blocks (Figure~\ref{fig:overview}). The Cutie object transformer, in particular, facilitates global communication between pixel-level features and object-level features without the use of computationally expensive attentional mechanisms such as cross-attention or self-attention. Instead, it integrates initial readouts, object queries, and object memory through transformer blocks, enabling bidirectional communication and enriching pixel features with object semantics to facilitate robust tracking across diverse scenarios. 

During the Cutie inference process, objects in each frame are segmented and labeled based on information drawn from a dynamically updated buffer of \textit{memory frames}.  Specifically, Cutie generates memory frames every $r$-th frame (where $r$ = the \textit{mem\_every} parameter, described below and in section~\ref{section:cutietracking}). The memory buffer comprises the initial, directly annotated frame (ground truth \textit{permanent memory}) plus the $T_{\max}$ most recently generated memory frames (managed via a First-In-First-Out (FIFO) strategy). Consequently, inference operations maintain consistent computational loads per frame and memory, independent of sequence length. Increasing $T_{\max}$ increases the size of the memory buffer and its computational cost while broadening the resources available for inference.  Increasing \textit{mem\_every} increases the speed of inference because fewer memory frames are generated, and also extends the distribution of the $T_{\max}$ memory frames maintained in the FIFO buffer further back and more sparsely in time.  The optimal value of \textit{mem\_every} from the perspective of accuracy alone is largely heuristic, and depends on factors such as the magnitude of movement per frame and the distribution of animal poses over time.  For simple tracking at standard frame rates, \textit{mem\_every} = 1 usually provides the greatest accuracy.

\subsection{Mask to Polygon Conversion}

\begin{figure*}[htbp]
    \centering
    \begin{minipage}{0.4\textwidth}
        \centering
        \includegraphics[width=6cm]{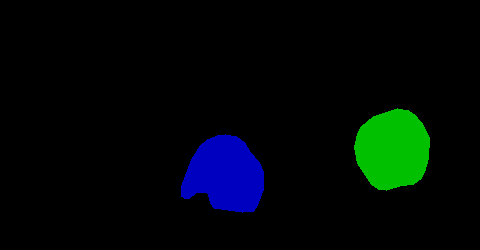}
        \caption*{Mask for Epsilon Value 1.0}
    \end{minipage}%
    \begin{minipage}{0.4\textwidth}
        \centering
        \includegraphics[width=6cm]{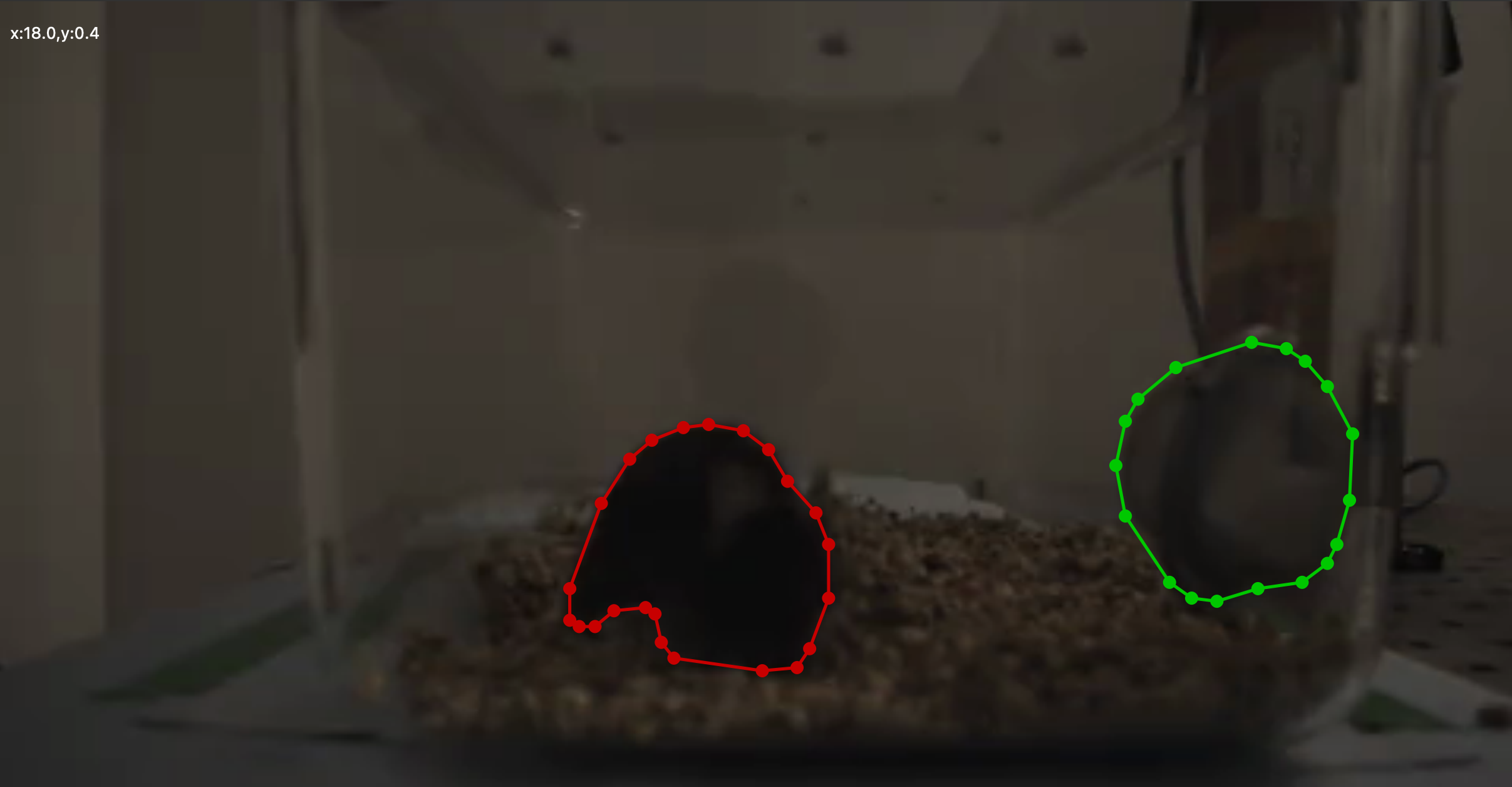}
        \caption*{Polygon for Epsilon Value 1.0}
    \end{minipage}\\%
    \begin{minipage}{0.4\textwidth}
        \centering
        \includegraphics[width=6cm]{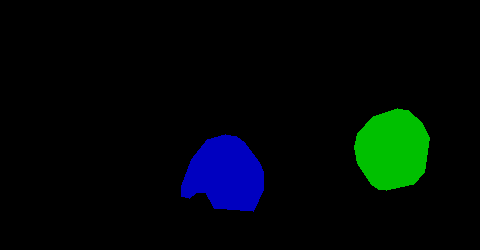}
        \caption*{Mask for Epsilon Value 2.0}
    \end{minipage}%
    \begin{minipage}{0.4\textwidth}
        \centering
        \includegraphics[width=6cm]{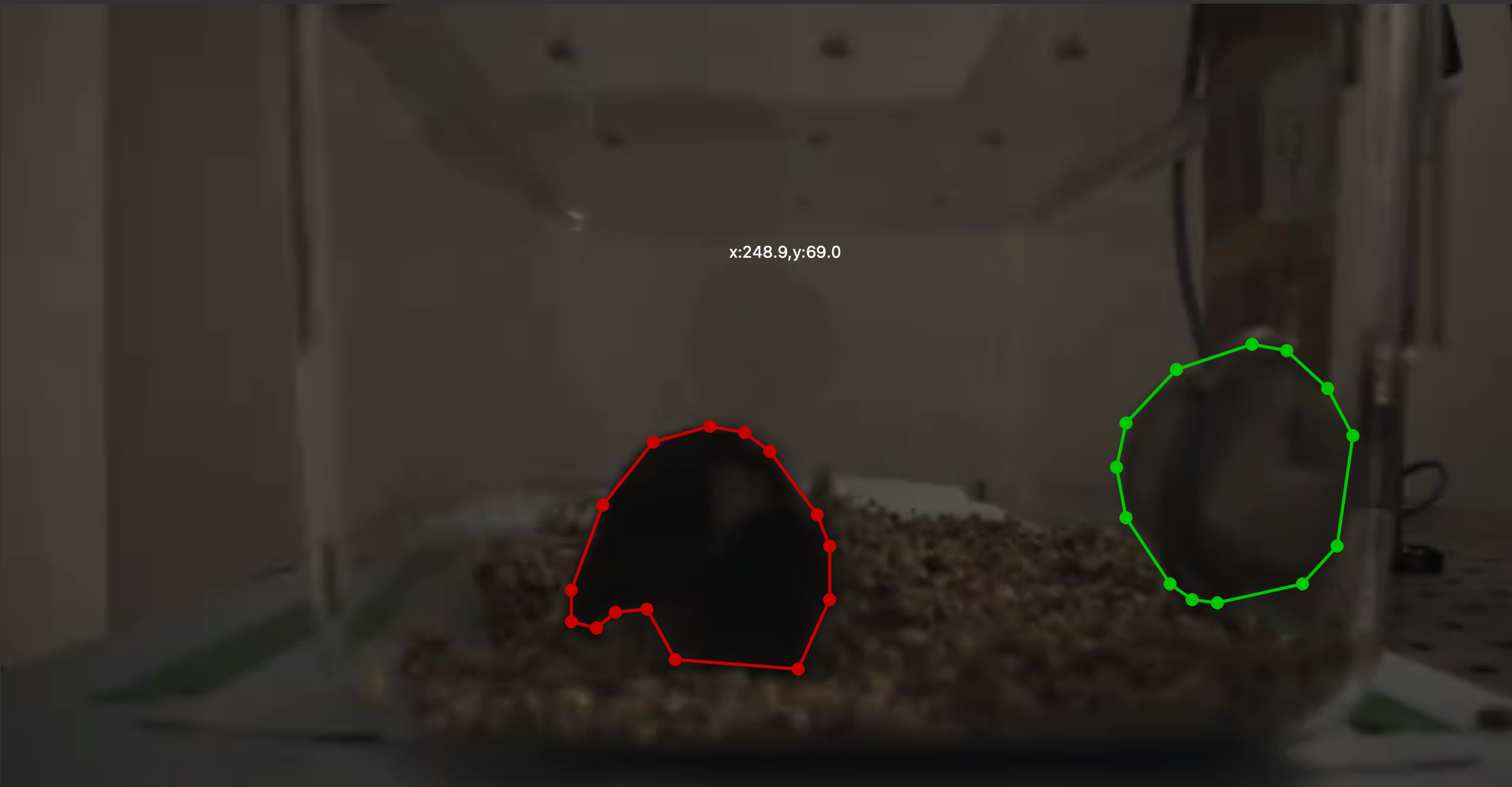}
        \caption*{Polygon for Epsilon Value 2.0}
    \end{minipage}\\%
    \begin{minipage}{0.4\textwidth}
        \centering
        \includegraphics[width=6cm]{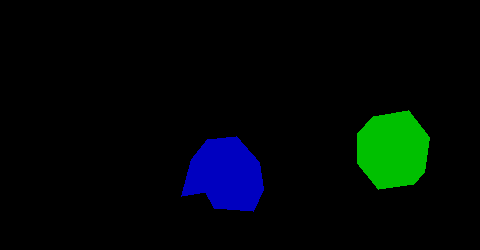}
        \caption*{Mask for Epsilon Value 4.0}
    \end{minipage}%
    \begin{minipage}{0.4\textwidth}
        \centering
        \includegraphics[width=6cm]{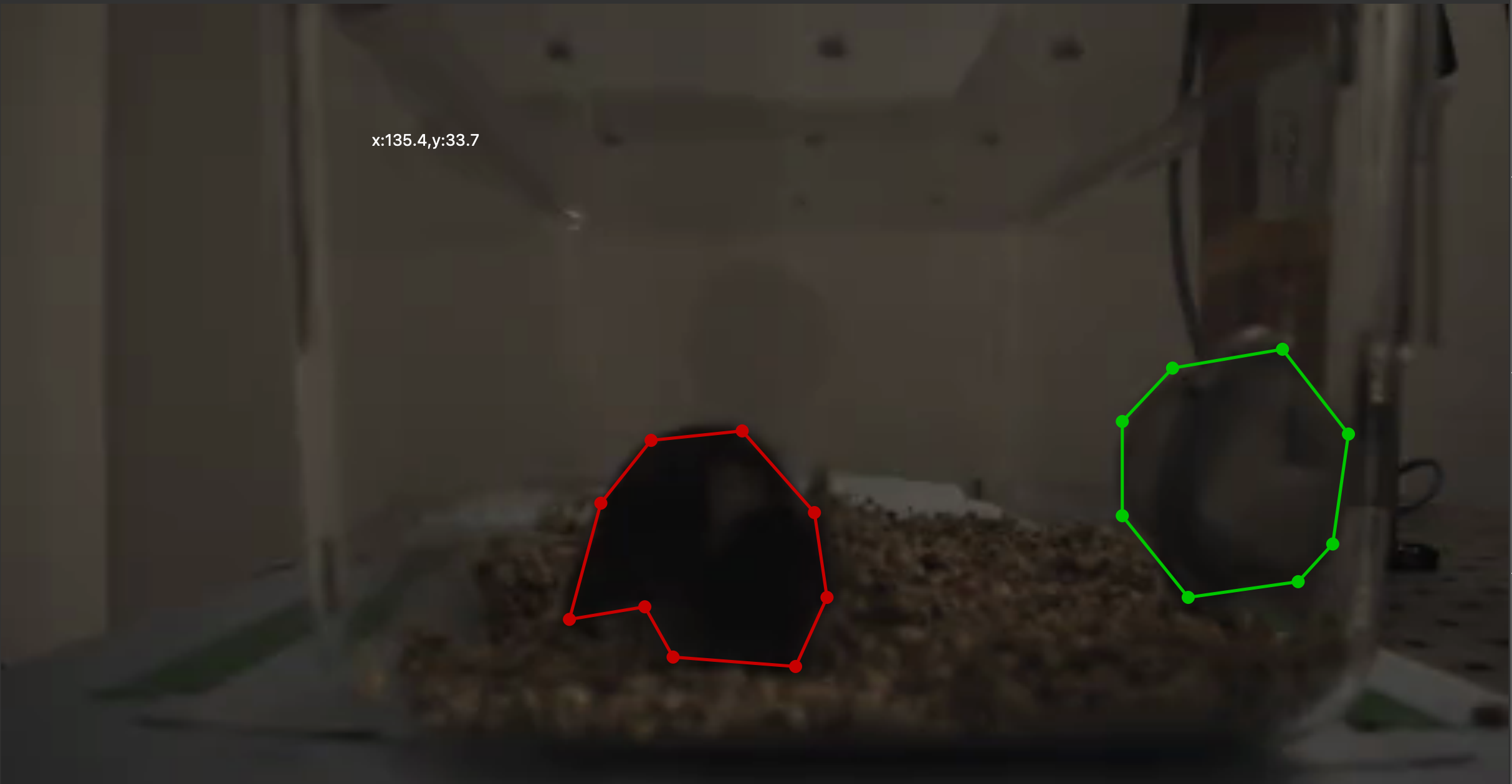}
        \caption*{Polygon for Epsilon Value 4.0}
    \end{minipage}\\%
    \begin{minipage}{0.4\textwidth}
        \centering
        \includegraphics[width=6cm]{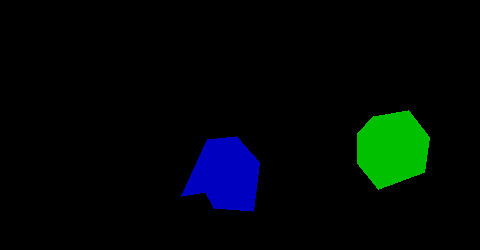}
        \caption*{Mask for Epsilon Value 8.0}
    \end{minipage}%
    \begin{minipage}{0.4\textwidth}
        \centering
        \includegraphics[width=6cm]{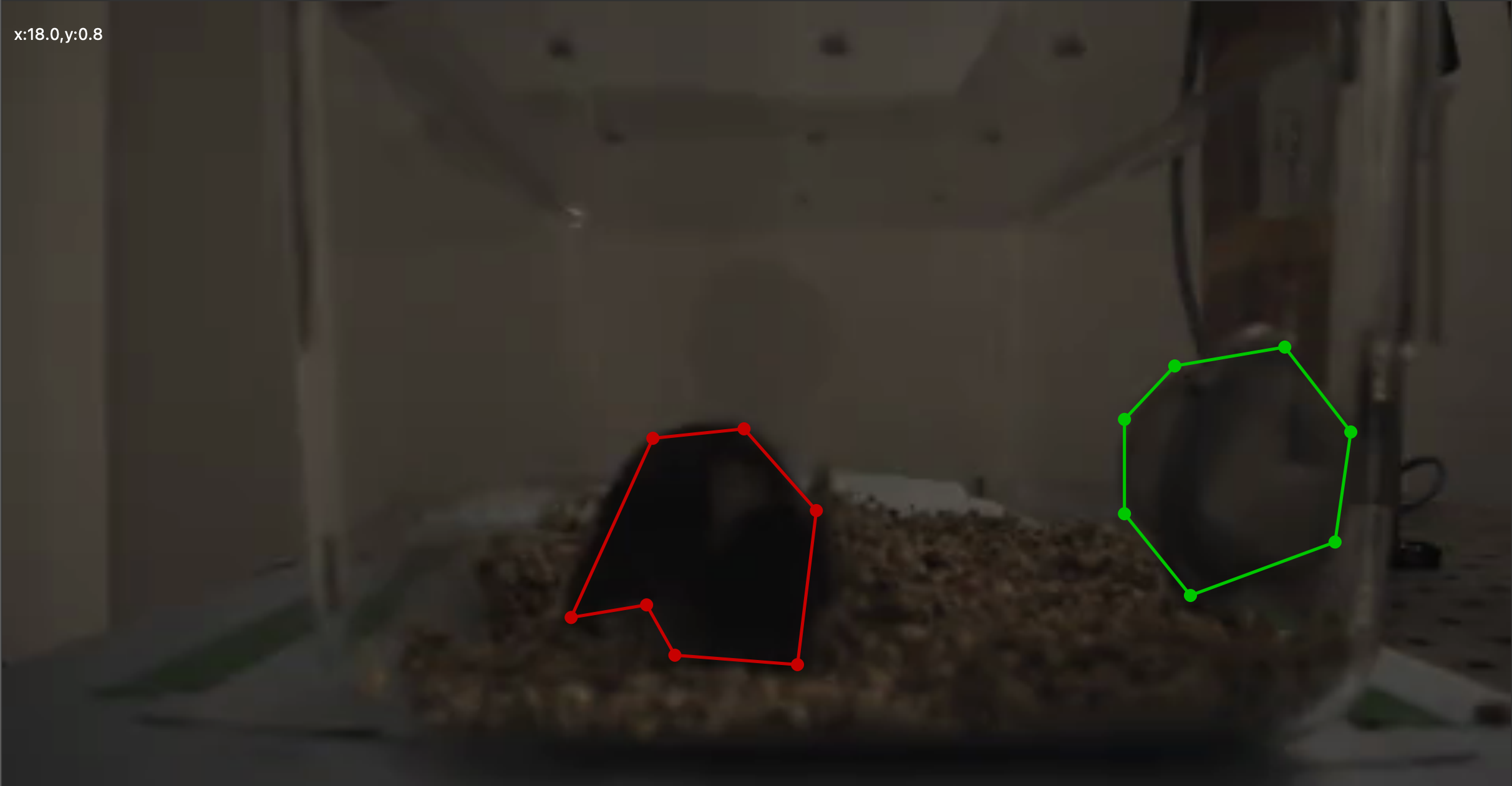}
        \caption*{Polygon for Epsilon Value 8.0}
    \end{minipage}
    \caption{Effects of different values of the epsilon parameter on the Annold mask-to-polygon converter. As epsilon is increased from 1 to 8, the polygons depicting each instance are generated with fewer points. An epsilon value of 2.0 typically preserves essential detail while limiting the number of vertices.}
    \label{fig:epsilon_masks_polygons}
\end{figure*}

In order to convert Cutie predictions back into editable Annolid polygons, we developed a mask-to-polygon conversion feature. A crucial parameter for this process is the optimization of polygon precision: specifying too few vertices generates crude and inaccurate masks, whereas excess vertices waste computational power and render polygons difficult to edit manually.  We specify a single parameter, $\epsilon$, to weight polygon precision; higher values of $\epsilon$ generate fewer points per polygon to depict each instance (Figure \ref{fig:epsilon_masks_polygons}).  After optimization testing using the cv2.approxPolyDP function of OpenCV \cite{opencv_library}, we determined that a value of $\epsilon = 2.0$ effectively balances smoothness and efficiency in shape representation, minimizing the number of vertices while preserving essential contour details. This default value can be altered via the \textit{Advanced Parameters} dialog accessible from the File menu of the Annolid GUI.

\subsection{Integrating Cutie into Annolid for Enhanced Tracking}
\label{section:cutietracking}

Annolid's method for multiple animal tracking uses a pretrained Cutie VOS model as the basis for the temporal propagation of segmentation predictions across sequential video frames. As described above, the VOS process begins with a single, fully annotated video frame identifying all relevant instances with polygon masks and corresponding labels.  The VOS model then predicts and segments objects in all subsequent video frames (end-to-end learning), eliminating the need to train a task-specific instance segmentation model. Frames that exhibit prediction errors can be identified during playback and corrected manually, after which the prediction process is re-engaged iteratively from the corrected frames. 

In Annolid, we have implemented several modifications and enhancements to the Cutie method to facilitate typical workflows in animal behavior analysis.  First, we updated the prediction method to recognize and log objective potential errors during inference, such as the loss of a tracked instance, and (optionally) to pause prediction automatically in response.  Specifically, if an existing instance is not detected in the current frame during Cutie prediction, as a fallback Annolid utilizes the most recent available bounding box for the missed instance and prompts SAM~\cite{kirillov2023segany} to predict the instance in that frame (this feature can be disabled in the \textit{Advanced Parameters} dialog, as it can be computationally expensive). If Annolid cannot predict or recover the instance with confidence, then the prediction will be automatically paused and the user prompted to correct the segmentation before continuing Cutie inference.  However, Annolid explicitly supports the case in which animals may disappear entirely from view and then later reappear (in which case they are automatically recognized and regain their previous labels).  Because genuinely hidden animals cannot in principle be distinguished from missed instances, this default auto-pause behavior may be undesirable. In such cases, users may elect to disable automatic pauses on error detection (in the Advanced Parameters dialog). Tracking errors in that case still will be logged, but will not pause prediction; any genuine errors then can be corrected after the fact, or via direct intervention during prediction as described below.  

Second, a live stop/restart prediction feature was added to facilitate real-time human interaction with the automatic prediction and segmentation process.  Once the user clicks the green \textit{Pred} (\textit{predict}) button in the GUI to initiate Cutie prediction, the button transforms into a red \textit{Stop} button while prediction is ongoing. By default, predictions for each frame are displayed as they are generated. (If the associated numeric spin box is updated during prediction, the step size for this video playback is altered but the \textit{mem\_every} parameter is not).  If the user clicks \textit{Stop}, prediction halts and the system navigates to the last predicted frame. Users then can navigate to any frame, correct or alter any annotations, and restart the prediction process from that point.  

Third, we have provided GUI access to an internal parameter termed \textit{mem\_every} (via a text spin box adjacent to the \textit{Open Video} button; Figure~\ref{fig:annolid_gui}). This parameter determines how often working memory is updated during processing (see Section~\ref{section:cutieprinciples}); higher values yield accelerated processing times whereas lower values yield more accurate predictions in typical datasets. That said, higher values of \textit{mem\_every} also can yield increased accuracy under certain circumstances.  

Finally, we have replaced the DEVA decoupled video segmentation approach proposed by the Cutie developers \cite{cheng2023tracking}, in which a separate image segmentation model is employed to detect and incorporate new instances entering the scene, with a method that intentionally does not segment any apparent new objects.  Instead, our approach assumes that all of the relevant ground truth instances are present and labeled in the initial designated frame, thereby eliminating the possibility of spurious false positive objects being identified in later frames. Existing objects that disappear and reappear are segmented and identified normally.

\subsection{The Annolid Annotation Framework}

The automatic segmentation, labeling, and predictive inference capacities offered by these new models supplement and enhance, but do not replace, the flexible, user-defined specification of arbitrary instances that provide Annolid with much of its versatility \cite{yang2023automated}. Automatic object segmentation and tracking as described herein coexist readily with trained Annolid instance segmentation models, and the respective outcomes can be measured and analyzed in combination.  Here, we focus on how the new models utilize Annolid's existing annotation mechanics, enabling human-in-the-loop use of Annolid annotation tools to manually correct or otherwise adjust segmentation masks in coordination with the new automated methods.  

The automatic object detection and segmentation process described above constructs detailed, editable polygons around each defined object from a specified class, each of which is assigned a unique label.  By default, the label nomenclature is to combine the class name with a sequential identification number (e.g., "fly\_13") to identify individual animals in the designated frame. The polygons delimiting each of these segmented objects are represented as specified points within a JSON file, making them accessible for manual editing using Annolid GUI tools. This foundational annotation of a designated frame sets the stage for tracking and analyzing multiple animals across the full video sequence.

\subsubsection{Segmentation and Prediction}

Following annotation, these polygons are transformed into binary instance masks for integration with the Cutie segmentation model. The Cutie model operates by reintroducing objects from its iteratively updated memory buffer into subsequent frames, thereby facilitating tracking. By merging Cutie's segmentation capabilities with precisely specified instance annotations, Annolid is able to deliver robust tracking performance across multiple animals within the video sequence.

\subsubsection{Manual Intervention and Correction}

After every epoch of automatic prediction, Cutie's predicted instance masks are converted back into polygons using Annolid's mask-to-polygon conversion feature.  This conversion enables researchers to visualize, assess, and manually edit tracked animal instances within Annolid's user interface. Specifically, users can visually observe the annotations arising from model predictions in each frame, either during the prediction process ("live") or after its completion.  If errors are observed, users navigate to the first inaccurate frame, edit or replace the polygons as desired to correct the annotation, and restart the automatic prediction process from this new designated frame. If desired, users can also choose to have objective errors, such as missing instances, pause the prediction process prematurely. If tracked animals are not expected to routinely vanish from view, this setting can save computational effort by directing users to fix such errors immediately after they occur.  

\subsubsection{Finalizing Predictions to Track Anything}

Manual corrections followed by automatic prediction based on the corrected frame can be iterated as often as necessary, and can be altered at any time. Users can adjust parameters such as \textit{mem\_every} before restarting automatic prediction in order to optimize accuracy and/or performance. Once the analyses of a given model are complete, Annolid can export tracking results in a human-readable CSV file format containing data fields such as instance name, frame number, centroid location (cx,cy), motion index in mask, bounding box, and segmentation mask encoded using COCO~\cite{lin2014microsoft} RLE format (Tables \ref{tab:motion_data}, \ref{tab:tracking_result_with_mask}). 

\section{Results}

We present our primary findings using a multiple animal tracking \& behavior (MATB) dataset previously assembled to test Annolid models \cite{yang2023automated} and a dataset of laboratory animal behavior videos assembled by the \textit{idTracker.ai} project \cite{romero2019idtracker}. The MATB dataset encompasses a wide array of tracking challenges, focusing on challenges such as partial occlusion during interactions, instances that vanish and reappear, camera motion, and reflections, as well as the identification of various behavioral states under these conditions. It includes both laboratory-generated videos and videos captured in the wild. The \textit{idTracker.ai} dataset includes laboratory animal scenarios featuring crowded environments with many animal interactions and brief moments of occlusion.  In each example, we utilized Annolid's Grounding DINO~\cite{liu2023grounding} and Segment Anything~\cite{kirillov2023segany} implementations to automatically annotate the first frame based on a (text) category label, and its Cutie VOS~\cite{cheng2023putting} implementation to predict segmentation and labeling across all of the remaining frames of the video.  Frames were not manually annotated (except for purposes of correcting model errors, as described below), and no traditional Annolid instance segmentation models \cite{yang2023automated} were trained in the process.  

\subsection{Evaluation with the MATB Dataset}

In this section, we demonstrate Annolid's multiple animal tracking performance using video data available from the Multiple Animal Tracking \& Behavior video compilation at ~\url{https://cplab.science/matb}. Individual video files are identified by their URLs. We evaluate the performance of Annolid's zero-shot learning, automatic segmentation, and tracking algorithms, including the number of human corrective interventions required, to assess their effectiveness and robustness under different scenarios.  Unless specified otherwise, the Annolid \textit{every\_mem} parameter was set to 1, $T_{max}$ = 5, and $\epsilon$ = 2.0.  

\subsubsection{Two mice}

In a video from the SLEAP project featuring paired male and female Swiss Webster mice \cite{Pereira2022sleap}, given the text prompt "mouse", Annolid automatically segmented and labeled both mice in the initial frame. This single annotated frame sufficed to successfully predict all instances across all 2559 frames of the video. To validate the tracking, we manually inspected the polygons overlaid on the video frames and found no instances of identity switches or tracking errors. The complete annotated video is available at \url{https://youtu.be/32vHPxiZpew}.

\subsubsection{Five goldfish}

In this original video of five goldfish in a tank, the fish repeatedly visually occluded one another and one fish fully exited and reentered the field of view.  Given the text prompt "fish", Annolid automatically segmented and labeled all five fish in the initial frame.  This annotated frame enabled automatic tracking of all fish through the video save for three errors requiring human correction. First, the fish\_0 instance was lost in frame 316. We backtracked to frame 277 to address a partial masking issue caused by occlusion and then resumed prediction from that point. Annolid paused again at frame 465 owing to the complete occlusion of fish\_4. Prediction was resumed after correction, and then paused again at frame 509 after fish\_4 had reappeared but subsequently was lost again due to reflections. After correction and resumption of the inference process, the fish were successfully tracked until the end of the video at frame 655. The annotated video is available at \url{https://youtu.be/CDtZ3efVlJU}. For comparison, we previously had labeled 50 frames in this video to train a traditional Annolid instance segmentation model~\cite{yang2023automated}. 

\subsection{Evaluation with the idTracker.ai Dataset}

In this section, we demonstrate Annolid's multiple animal tracking performance using video data available from the \textit{idTracker.ai} project (a Google Drive-based data repository linked from \url{https://idtracker.ai}). Individual video files are identified below by their filenames in that repository. We evaluate the performance of Annolid's zero-shot learning, automatic segmentation, and tracking algorithms, including the number of human corrective interventions required, to assess their effectiveness and robustness under different scenarios.

The videos within the \textit{idTracker.ai} dataset are presented and validated at frame rates from 30 to 100 frames per second (fps), and encoded at a variety of resolutions; many exceed 10 minutes in length (i.e., 18,000 frames for a 30 fps video). We resized the test videos to lower resolutions for computational efficiency, as detailed in each section below, but maintained the original frame rates and video durations.  Unless specified otherwise, the Annolid \textit{every\_mem} parameter was set to 1, $T_{max}$ = 5, and $\epsilon$ = 2.0.

\subsubsection{Two Mice}

We employed ffmpeg to compress the 'mice\_2\_2.avi' video file using the command 'ffmpeg -i mice\_2\_2.avi -vcodec libx264 -vf scale=984:557'. This compression reduced the file size from 12 gigabytes to approximately 100 megabytes in MP4 format. Next, we entered the text prompt 'mouse' in the Annolid GUI, invoking the Grounding DINO and SAM-HQ models to generate initial frame polygons for both the 'mouse\_1' and 'mouse\_2' instances. These polygons then were applied via the Cutie method to track all 23,520 frames of the video, generating a tracking accuracy of 100\% without any missing instances or identity switching. The complete annotated video is available for viewing at \url{https://youtu.be/lIPk92bOMxw}.

\subsubsection{Four mice}

We next used Annolid to track four mice in the video \textit{mice\_4\_1.avi}, first compressing the 17 GB file to 173 MB using the libx264 codec as above with \textit{scale=1272:909}. Given the text prompt "mouse," Annolid successfully generated polygons for each of the 4 mice in the first frame (mouse\_0 through mouse\_3; Figure~\ref{fig:teaser}). Again, this single annotated frame sufficed to process the entire ten-minute video, comprising 15,769 frames. To validate the tracking, we manually inspected the polygons overlaid on the video frames and found no instances of identity switches or tracking errors. The complete annotated video is available for viewing at ~\url{https://youtu.be/PNbPA649r78}.
 
\subsubsection{Fourteen ants}

We then used Annolid to track 14 ants in the video \textit{ants\_14.MOV}, first compressing the 3.1 GB file to 105 MB using the libx264 codec with \textit{scale=1280:720}. Given the text prompt "ant", Annolid successfully labeled all 14 instances of ants in the first frame (ant\_0 through ant\_13; Figure~\ref{fig:teaser}).  Again, this sufficed to predict instances successfully throughout the entire video, comprising 35,964 frames. To validate the tracking, we manually inspected the polygons overlaid on the video frames and found no instances of identity switches or tracking errors. The complete annotated video is available for viewing at ~\url{https://youtu.be/iqhz1R79EZg}.

\subsubsection{Six \textit{Drosophila} fruit flies}

We then used Annolid to track six \textit{Drosophila} fruit flies in the video \textit{drosophila\_6.avi}, first compressing the 15.5 GB file to 48.9 MB using the libx264 codec with \textit{scale=804:808}. Given the text prompt "drosophila", Annolid successfully labeled all six instances of flies in the first frame.  However, unlike the examples above, Annolid made three errors during Cutie inference that required human-in-the-loop editing.  Specifically, predictions were automatically paused for corrections at the following three points in the video:

\begin{itemize}
    \item Prediction halted at frame 10043 due to a polygon issue concerning \texttt{drosophila\_0}. The polygon was manually edited using Annolid GUI tools and prediction was restarted.  
    \item Similarly, at frame 10188, prediction halted owing to another minor polygon issue with \texttt{drosophila\_0}. The polygon was edited and prediction was restarted.
    \item At frame 16045, \texttt{drosophila\_0} was reported missing, prompting a backtrack to frame 15975 to rectify a polygon problem. The polygon was edited and prediction was restarted. Prediction subsequently was successful through the end of the video (frame 25579).  
\end{itemize}

\begin{figure*}[t]
  \centering
  \begin{flushleft}
\begin{tabular}{c@{}c@{\hspace{-1mm}}c@{\hspace{-1mm}}c}

    \rotatebox[origin=c]{90}{\small Images \hspace{1mm}}&
    \raisebox{-0.5\height}{%
        \frame{\includegraphics[trim={0.03cm 0 0.03cm 0},clip,width=0.30\linewidth]{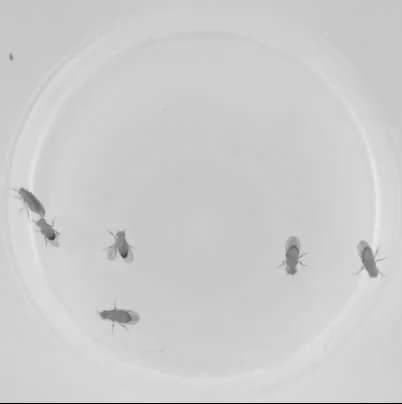}}
    }&
    \raisebox{-0.5\height}{%
        \frame{\includegraphics[trim={0.03cm 0 0.03cm 0},clip,width=0.30\linewidth]{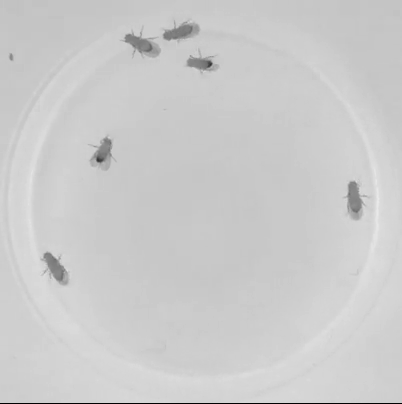}}
    }&
    \raisebox{-0.5\height}{%
        \frame{\includegraphics[trim={0.03cm 0 0.03cm 0},clip,width=0.30\linewidth]{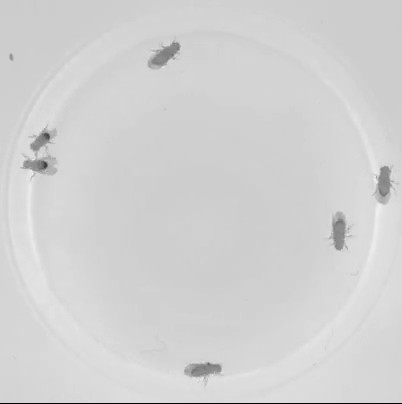}}
    }\\
    \vspace{-0.03mm}\\
    \rotatebox[origin=c]{90}{\small Polygons \hspace{1mm}}&
    \raisebox{-0.5\height}{%
        \frame{\includegraphics[trim={0.03cm 0 0.03cm 0},clip,width=0.30\linewidth]{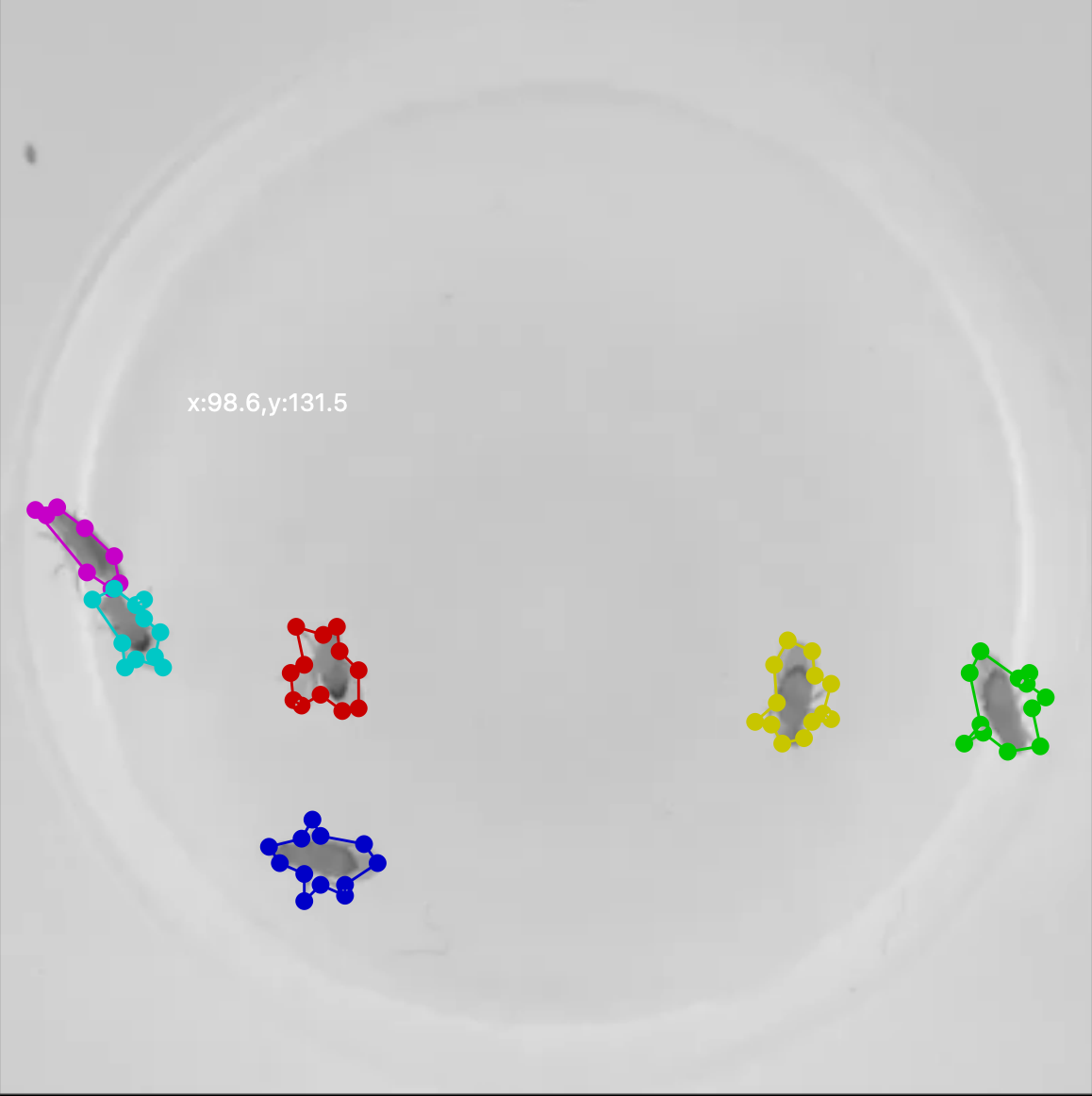}}
    }&
    \raisebox{-0.5\height}{%
        \frame{\includegraphics[trim={0.03cm 0 0.03cm 0},clip,width=0.30\linewidth]{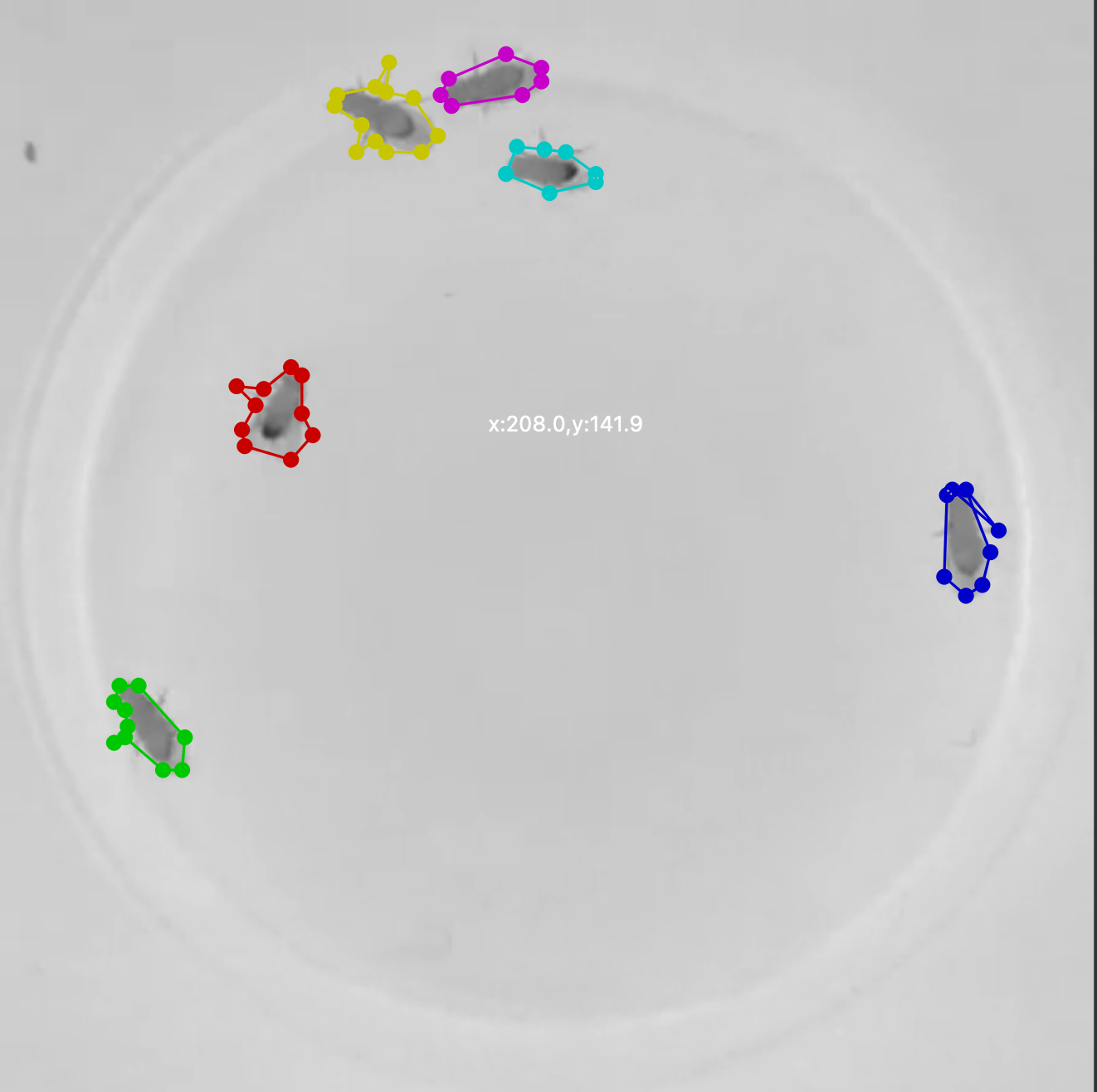}}
    }&
    \raisebox{-0.5\height}{%
        \frame{\includegraphics[trim={0.03cm 0 0.03cm 0},clip,width=0.30\linewidth]{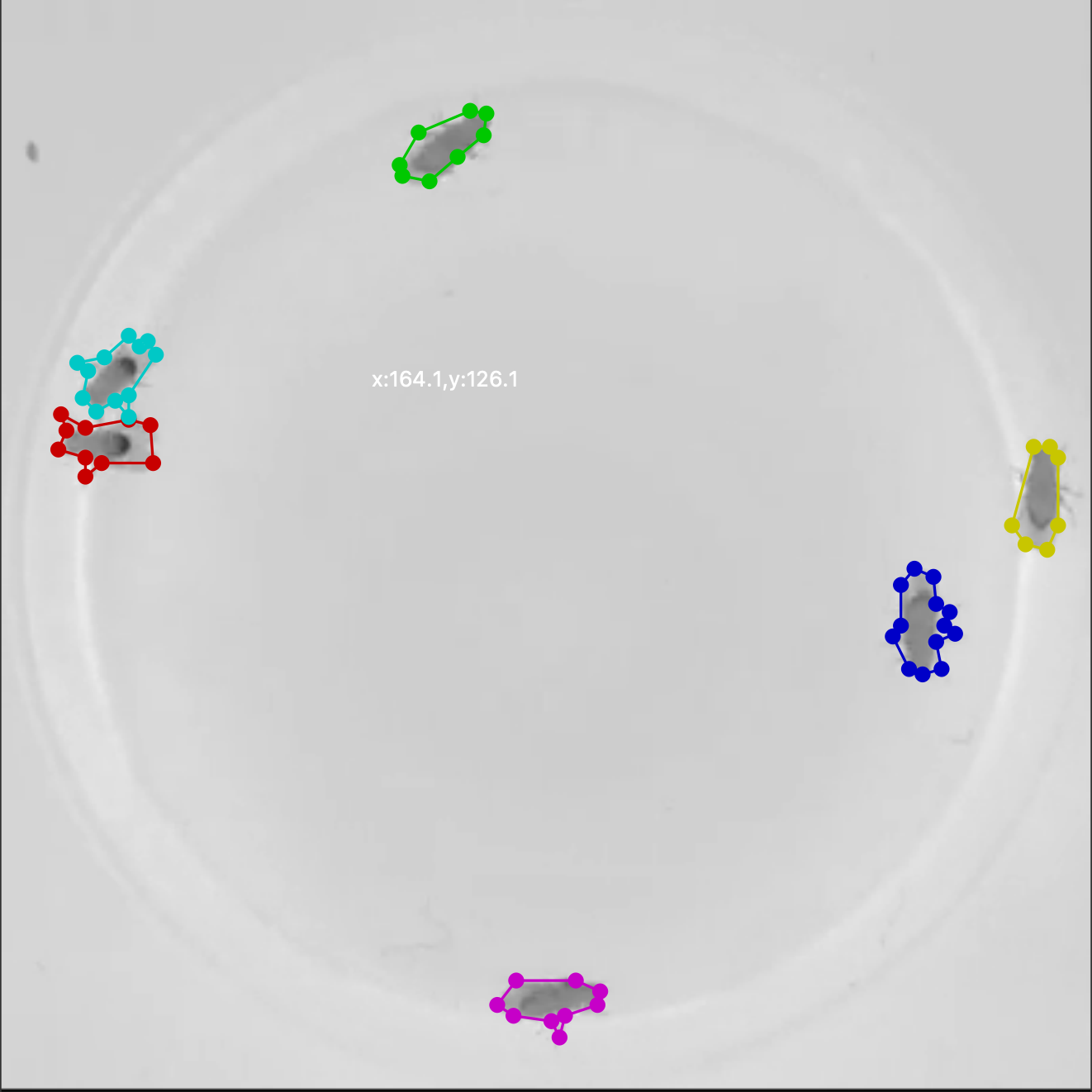}}
    }\\
    \vspace{-0.03mm}\\
    
    \end{tabular}
\end{flushleft}
  \caption{Annolid performance on an idTracker.ai video \cite{romero2019idtracker} featuring markerless tracking of six \textit{Drosophila} fruit flies in an arena. From left to right: frames \#1, \#2000, and \#4000 are shown. The complete video is available at \url{https://youtu.be/uTs6CKgmdSw}.}
  \label{fig:drosophila-vis}
\end{figure*}

Instances where polygons briefly enlarged and then returned to an appropriate size did not prompt intervention. To validate the final tracking, we manually inspected the polygons overlaid on the video frames and found no instances of ID switches or missed tracking. An example frame in the Annolid GUI is depicted in Figure~\ref{fig:drosophila-vis}; the complete annotated video is available for viewing at ~\url{https://youtu.be/uTs6CKgmdSw}.

\subsubsection{Seven zebrafish}
\label{section:sevenzebrafish}

We then used Annolid to track the seven zebrafish in the video \textit{zebrafish\_7.avi}, first compressing the 17.6 GB file to 59.2 MB using the libx264 codec with \textit{scale=992:998}. Given the text prompt "fish", Annolid successfully labeled all seven instances of zebrafish in the first frame (Figure~\ref{fig:teaser}).  However, predictions of subsequent frames proved more challenging in this video file. Even with the \textit{mem\_every} parameter set to 1 (which offered maximum accuracy in this context), automatic prediction from the start of the file only tracked all fish accurately for 447 frames (Table~\ref{tab:test_mem_every}).  

To track all seven fish throughout the full 10-minute (19,079 frames) video, we applied Annolid's (default) option to pause prediction once tracking failed for at least one instance. At each of these pauses, we manually corrected and identified the missing instance using point prompting and initiated the prediction process again. This process was repeated until the entire video was fully tracked.  Ultimately, 24 iterations were required, yielding an error rate of 0.1256\%, or less than 1.3 errors per thousand video frames.  The complete annotated video is available for viewing at ~\url{https://youtu.be/adeHXZEaYXQ}. 

We attribute the relative difficulty of automatic frame prediction in this 31.36 fps video file to the fast motion and sudden turns made by the fish, which led to increased frame-by-frame differences in the features of each instance.  We expect that recording at higher frame rates would improve the performance of automatic prediction by reducing the magnitude of these effects.

\begin{table}[htbp]
     \centering
     \small 
     \begin{tabular}{@{}ll@{}}
       \toprule
       \textbf{\textit{mem\_every}} & \textbf{Frames Tracked} \\
       \midrule
       1 & 447 \\
       2 & 447 \\
       3 & 325 \\
       5 & 6 \\
       \bottomrule
     \end{tabular}
     \caption{Tracking performance as a function of \textit{mem\_every} on the video described in Section~\ref{section:sevenzebrafish}, \textit{Seven zebrafish}.  Performance here is operationalized as the number of frames tracked without error starting at the beginning of the video file.  \textit{Mem\_every} values of 1 cause every frame to be encoded as a memory frame and pushed into the $T_{max}$-element FIFO memory buffer for use in Cutie prediction.  Values of 2 cause every second frame to be encoded as a memory frame, values of 3 every third frame, etcetera. See Section~\ref{section:cutieprinciples} for details.}
     \label{tab:test_mem_every}
\end{table}

\subsubsection{Ten \textit{Drosophila} fruit flies}

We then used Annolid to track ten \textit{Drosophila} fruit flies in the ten-minute, 60 fps video \textit{drosophila\_10.avi}, first compressing the 355 MB file to 8.6 MB using the libx264 codec with \textit{scale=640:480}. Given the text prompt "fly", Annolid successfully labeled all ten instances of flies in the first frame.  During subsequent Cutie inference, Annolid processed without error up to frame 18388. After we backtracked to frame 18386, restored the missing polygon for fly\_2, and restarted inference, Annolid successfully predicted through to the end of the video (frame 36769).  To validate the final tracking, we manually inspected the polygons overlaid on the video frames and found no instances of identity switches or tracking errors. The complete annotated video is available for viewing at ~\url{https://youtu.be/23Qtm9esxF8}.

\subsubsection{Eighty \textit{Drosophila} fruit flies}
\label{section:eightyflies}

We then used Annolid to track eighty \textit{Drosophila} fruit flies in the video \textit{drosophila\_80.avi}, first compressing the 19.7 GB file to 15.9 MB using the libx264 codec with \textit{scale=928:900}. Text prompting did not successfully segment flies in the first frame. This fully automatic segmentation might have succeeded at the original video resolution, but the computational costs and memory load required for processing at that resolution would increase considerably.  Instead, it proved more efficient to use Annolid's SAM-based point prompting method to annotate the initial frame; we clicked on each fly in turn and Annolid outlined and segmented the instance automatically.  

During subsequent Cutie inference, over the 20,494 frames of the video, Annolid encountered 292 total errors (error rate of 1.42\%), which were iteratively corrected as described above.  (The errors included missing fly instances and polygon imperfections, exacerbated by the tendency of flies to suddenly take flight and move large distances between frames, and are detailed in Table~\ref{tab:prediction_issues_combined}). To validate the final tracking result, we manually inspected the polygons overlaid on the video frames and found no instances of identity switches or missing instances.  The complete annotated video is available for viewing at \url{https://youtu.be/D50RYbBR8Ho}.

The Cutie model's error rate was higher than for the other examples presented herein, presumably owing to the large number of instances together with their visual similarity (exacerbated by the limited resolution of the compressed video) and repeated close proximity to one another.  Correcting these errors and verifying the tracking accuracy with this sample of 80 fruit flies proved highly time-consuming, in contrast to the other examples discussed herein, but may still compare favorably to other methods if the accurate tracking of all 80 flies is an end-user goal. This scenario presents a challenge for future improvements in model efficiency. 

\section{Discussion}

Several existing approaches in multiple animal tracking utilize pose estimation networks trained on annotated video datasets;  notable examples include DeepLabCut~\cite{Lauer2022MultianimalPE} and SLEAP~\cite{Pereira2022sleap}. Alternatively, idTracker~\cite{perez2014idtracker} and idTracker.ai~\cite{romero2019idtracker} employ threshold-based segmentation to divide animals into distinct blobs for tracking purposes. In the case of idTracker.ai, the occlusion problem is addressed by utilizing neural networks to classify segmented blobs as either \textit{crossing} or \textit{non-crossing}, with a separate network dedicated to identifying individual animal identities.  Other researchers have extensively explored the concept of 'tracking-by-detection'~\cite{kim2015multiple,tang2017multiple,bergmann2019tracking}. While these approaches differ from one another, each of them typically treat image-level detections as fixed entities, with the temporal model focusing solely on associating these detected objects. This formulation relies heavily on the accuracy of per-image detections and is susceptible to errors at the image level.  Accordingly, the scalability of these packages to handle large-vocabulary or open-world video data is unclear. As the number of instances and scenarios in a dataset multiply, the task of training and developing end-to-end models to jointly tackle pose estimation, segmentation, and association becomes increasingly daunting, particularly when annotations are sparse.

To the best of our knowledge, recent advancements in video segmentation methods~\cite{cheng2022xmem,cheng2023putting}, particularly those tailored for open-world settings such as BURST~\cite{athar2023burst}, have primarily been trained end-to-end. Our reliance on the Cutie pretrained model, which demonstrates adaptability across various tasks, underscores the efficacy of the present approach. Recent strides in universal promptable image segmentation models, including the SAM model used herein~\cite{kirillov2023segany} along with several others~\cite{zou2023segment,li2023semantic,sam_hq,mobile_sam,zhao2023fast,xiong2023efficientsam,cai2022efficientvit}, offer promising alternatives to image-level models, enabling the labeling of all instances in a frame with text or point-based prompts.

\section{Conclusion}
In this study, we introduce a coordinated set of new machine learning methods into Annolid and leverage these tools to greatly improve the efficiency and accuracy of multi-animal tracking analysis.  Specifically, we introduce a Cutie-based prediction strategy for multiple instance tracking in video sequences based on single-frame annotations, a SAM-based strategy for automatic segmentation and labeling of instances in this initial frame, and a Grounding DINO-based strategy for specifying the objects to segment and label by name, via text prompt.  Multiple animals therefore can be tracked throughout entire research videos without manually designating or annotating a single one.  Additionally, we update the Annolid toolset and GUI to facilitate the easy validation and manual correction of automatic tracking results.  This combination of updates offers a streamlined approach that can replace traditional methods reliant on manually labeled datasets for fine-tuning neural network training. Configuration options enable users to strike a balance between tracking accuracy and computational efficiency, thereby laying the groundwork for real-world deployment in animal behavior analysis and biomedical research applications. 

Multi-animal tracking is an important and widespread requirement for many applications in animal behavior analysis, but there are many other applications and research goals to which the advances described herein do not directly apply.  For such applications, Annolid retains its core instance segmentation strategy~\cite{yang2023automated}, in which users annotate some number of frames with behaviors or other features of interest and train Annolid models on this training set. Some of the present advances, such as the use of point prompting to automatically segment selected instances with SAM, can be applied toward these ends -- e.g., to more easily identify animal body configurations signifying a particular behavioral state or pose.  Other applications may require manual polygon annotation.  Overall, Annolid enables users to combine these diverse analytical approaches in order to design the most appropriate and efficient strategy with which to address their own particular research questions.  

\vspace{0.2in}
{
\small
\noindent\textbf{Acknowledgments}. This work was supported by NIH/NIDCD grants R01 DC019124 and R01 DC014701.
}
\vspace{0.3in}

{\small
\bibliographystyle{IEEEtran}
\bibliography{main}
}

\clearpage
\appendix
\beginsupplement
\onecolumn

\section{Supplementary Data}

\begin{table}[htbp]
    \centering
    \caption{Example of the Annolid Tracked CSV format output, featuring columns for Frame Number, Instance Name, cx, cy, Motion Index, and Timestamps. (cx, cy) are the centroid coordinates of the instance mask, measured in pixel units. The Motion Index is computed as the sum of magnitudes of the polar coordinates of the optical flow overlaid on the instance mask, divided by the mask size \cite{yang2023automated}. The naming convention for this file follows the pattern '\textit{videoname}\_tracked.csv'.}
    \label{tab:motion_data}
    \begin{tabular}{cccccc}
        \toprule
        Frame Number & Instance Name & \(cx\) & \(cy\) & Motion Index & Timestamps \\
        \midrule
        0 & mouse & 324 & 141 & -1.0 & 00:00:00 \\
        0 & teaball & 99 & 236 & -1.0 & 00:00:00 \\
        1 & mouse & 326 & 146 & 1.099 & 00:00:00.024 \\
        1 & teaball & 98 & 236 & 0.0 & 00:00:00.024 \\
        2 & mouse & 329 & 150 & 1.112 & 00:00:00.048 \\
        2 & teaball & 98 & 236 & 0.0 & 00:00:00.048 \\
        3 & mouse & 331 & 153 & 1.256 & 00:00:00.072 \\
        3 & teaball & 98 & 236 & 0.0 & 00:00:00.072 \\
        4 & mouse & 333 & 157 & 1.076 & 00:00:00.096 \\
        4 & teaball & 98 & 236 & 0.0 & 00:00:00.096 \\
        5 & mouse & 335 & 159 & 0.817 & 00:00:00.120 \\
        5 & teaball & 98 & 236 & 0.0 & 00:00:00.120 \\
        6 & mouse & 336 & 162 & 0.655 & 00:00:00.144 \\
        6 & teaball & 98 & 236 & 0.0 & 00:00:00.144 \\
        7 & mouse & 337 & 165 & 0.769 & 00:00:00.168 \\
        7 & teaball & 98 & 236 & 0.0 & 00:00:00.168 \\
        8 & mouse & 339 & 168 & 0.927 & 00:00:00.192 \\
        \bottomrule
    \end{tabular}
\end{table}

\begin{table*}[htbp]
    \centering
    \caption{Predictions for the tracking data of a video featuring a mouse interacting with a teaball, presented in a CSV file format generated by Annolid.  The output includes the frame number, bounding box coordinates $x1$, $y1$, $x2$, $y2$, center coordinates $cx$, $cy$, and the segmentation mask encoded in  COCO format. The naming convention for this CSV file follows the pattern '\textit{videoname}\_tracking.csv'. When this file is placed in the same directory as the corresponding video file, Annolid automatically loads the predicted instances in polygons or points, along with other shapes, from this CSV file.  This facilitates quality control and editing during video analysis.}
    \label{tab:tracking_result_with_mask}
    \begin{threeparttable}
    \begin{tabular}{|c|c|c|c|c|c|c|c|c|}
    \hline
    \textbf{Frame} & \textbf{x1} & \textbf{y1} & \textbf{x2} & \textbf{y2} & \textbf{cx} & \textbf{cy} & \textbf{Instance Name} & \textbf{Segmentation} \\ \hline
    0 & 117.0 & 148.0 & 175.0 & 227.0 & 143.74 & 185.36 &  mouse1040  & \{'``[13U94......MmoY2'\\ \hline
    0 & 212.0 & 70.0 & 263.0 & 128.0 & 236.88 & 100.41 &  teaball  & \{'ngd01......HllV3'\\ \hline
    1 & 116.0 & 150.0 & 174.0 & 225.0 &143.63 & 185.32 &  mouse1040  & \{'XS\textbackslash 14......IobZ2'\\ \hline
    1 & 212.0 & 70.0 & 263.0 & 128.0 & 236.72 & 100.64 & teaball  & \{'hgd01......MPmV3'\\ \hline
    2 & 117.0 & 149.0 & 175.0 & 226.0 & 143.93 & 185.77 & mouse1040  & \{ 'li[12V94......L]YZ2'\\ \hline
    2 & 212.0 & 70.0 & 263.0 & 130.0 & 236.73 &  101.40 & teaball  & \{'hgd04......JSZV3'\\ \hline
    3 & 117.0 & 148.0 & 175.0 & 226.0 & 143.91 & 185.66 & mouse1040  & \{'``[12W93......K\textbackslash YZ2'\\ \hline
    3 & 212.0 & 70.0 & 263.0 & 130.0 & 236.85 & 102.36 & teaball  & \{'hgd01......MkPV3'\\ \hline
    4 & 118.0 & 148.0 & 175.0 & 226.0 & 143.50 & 185.90 & mouse1040  & \{'``[12X92......1Oj\^[2'\\ \hline  
    4 & 214.0 & 70.0 & 263.0 & 131.0 & 236.84 & 102.31 & teaball  & \{'hgd01......IgPV3'\\ \hline
    5 & 118.0 & 148.0 & 175.0 & 227.0 & 143.64 & 185.84 & mouse1040  & \{'``[12X92......NooY2'\\ \hline
    5 & 214.0 & 70.0 & 263.0 & 131.0 & 236.85 & 102.36 & teaball  & \{'hgd01......MjPV3'\\ \hline
    \end{tabular}
    \begin{tablenotes}
    \item[a] Note that the middle section of the segmentation COCO format encoding has been replaced with dots for compact display purposes.  The segmentation column in the tracking CSV file comprises the complete encoding, consistently commencing with "'size':['H','W'],'counts':". In this encoding, 'size' denotes the video's dimensions in pixel units, encompassing its height H and width W, whereas the 'counts' value corresponds to the COCO format Run-Length Encoding (RLE).
    \end{tablenotes}
    \end{threeparttable}
\end{table*}

\begin{longtable}{|c|p{0.25\linewidth}|p{0.55\linewidth}|}
    \caption{Listing of issues and corrections made during prediction for the video \textit{drosophila\_80.mp4}. See section ~\ref{section:eightyflies} for details.}
    \label{tab:prediction_issues_combined} \\
    \hline
    \textbf{Frame Number} & \textbf{Issue Encountered} & \textbf{Actions Taken} \\
    \hline
    \endfirsthead
    
    \multicolumn{3}{c}%
    {{\bfseries \tablename\ \thetable{} -- continued from previous page}} \\
    \hline
    \textbf{Frame Number} & \textbf{Issue Encountered} & \textbf{Actions Taken} \\
    \hline
    \endhead
    
    \hline \multicolumn{3}{|r|}{{Continued on next page}} \\ \hline
    \endfoot
    
    \hline \hline
    \endlastfoot
    333 & Tracking lost for 'fly32'. & Reverted back to frame 298 and identified an imperfect polygon. Ceased prediction. Rectified the issue before resuming prediction from frame 334. \\
    349 & Missing 'fly34'. & Resolved at 349, then resumed prediction. \\
    393 & Missing 'fly32'. & Resolved at 393, then resumed prediction.\\
    681 & Enlarged polygon of 'fly27' and 'fly31'.& Resolved at 681, then resumed prediction. \\
    881 & Corrected polygon for 'fly5'. & Resolved at 881, then resumed prediction. \\
    1000 & Corrected polygon for 'fly19'. & Resolved at 1000, then resumed prediction.\\
    1203, 1238, 1277 & Corrected 'fly39'. & Resolved at 1203, 1238, 1277, then resumed prediction.\\
    1448 & Missing 'fly31'.  & Resolved at 1448, then resumed prediction.\\
    1555 & Fixed polygons for 'fly21', 'fly28', and 'fly66'. & Resolved at 1555, then resumed prediction. \\
    1700 & Fixed polygon for 'fly56'.  & Resolved at 1700, then resumed prediction. \\
    1899 & Missing 'fly39'.  & Resolved at 1899, then resumed prediction.\\
    2120 & Missing 'fly55'.  & Resolved at 2120, then resumed prediction.\\
    2278 & Black dot on 'fly32' polygon & Corrected polygon for 'fly32' due to the black dot \\
    2419 & Missing 'fly31' & Corrected polygon for 'fly17'; handled missing 'fly31' \\
    2587 & Enlarged 'fly65' polygon & Corrected polygon size for 'fly65' \\
    3256 & Missing 'fly31'; spiking 'fly32' & Corrected 'fly32' polygon; handled missing 'fly31' with large motion \\
    3701 & Missing 'fly32'; spiking 'fly79' & Handled missing 'fly32'; corrected 'fly79' polygon \\
    3871 & Resized 'fly18' polygon & Resized polygon for 'fly18' \\
    3887 & Missing 'fly54'; spiking 'fly79' & Handled missing 'fly54' with large motion; corrected 'fly79' polygon \\
    4039 & Missing 'fly0' flying & Addressed missing 'fly0' flying \\
    4045 & Missing 'fly12' & Addressed missing 'fly12' \\
    4200 & Enlarged polygon for 'fly66' & Fixed enlarged polygon for 'fly66' \\
    4453 & Missing 'fly79' & Addressed missing 'fly79' \\
    4563 & Missing 'fly66' & Addressed missing 'fly66' \\
    4704 & Missing 'fly77'; Missing 'fly40' & Addressed missing 'fly77'; Addressed missing 'fly40' \\
    4908 & Missing 'fly21'; Spiking polygons for 'fly79' & Addressed missing 'fly21'; Addressed spiking polygons for 'fly79' \\
    5126 & Enlarged 'fly49' polygon & Fixed the enlarged polygon for 'fly49' \\
    5148 & Missing 'fly34' & Addressed missing 'fly34' \\
    5203 & Enlarged 'fly49' polygon & Fixed the enlarged polygon for 'fly49' \\
    5246 & Missing 'fly66' & Addressed missing 'fly66' \\
    5258 & Missing 'fly74' flying & Addressed missing 'fly74' flying \\
    5424 & Missing 'fly63' flying & Addressed missing 'fly63' flying \\
    5471 & Missing 'fly79' & Addressed missing 'fly79' \\
    5554 & Missing 'fly31' & Addressed missing 'fly31' \\
    5824 & Enlarged 'fly78' polygon & Fixed the enlarged polygon for 'fly78' \\
    5887 & Enlarged 'fly32' polygon & Fixed the enlarged polygon for 'fly32' \\
    5929 & Prediction start & Checked this frame \\
    6029 & Missing 'fly32' & Resolved at 6014, then resumed prediction. \\
    6379 & Missing 'fly19' & Resolved at 6379, then resumed prediction. \\
    6994 & Missing 'fly77' & Resolved at 6993, then resumed prediction. \\
    7045 & Missing 'fly22' & Resolved at 7042, then resumed prediction. \\
    7114 & Missing 'fly16' & Resolved at 7113, then resumed prediction. \\
    7241 & Missing 'fly8' & Resolved at 7239, then resumed prediction. \\
    7114 & Missing 'fly16' & Resolved at 7114, then resumed prediction. \\
    7633 & Missing 'fly55' & Resolved at 7613, then resumed prediction. \\
    7624 & Missing 'fly32' & Resolved at 7623, then resumed prediction. \\
    7653 & Missing 'fly37' & Resolved at 7652, then resumed prediction. \\
    7654 & Missing 'fly77' & Resolved at 7648, then resumed prediction. \\
    7661 & Missing 'fly0' & Resolved at 7661, then resumed prediction. \\
    7700 & Missing 'fly79' & Resolved at 7699, then resumed prediction. \\
    7701 & Missing 'fly76' & Resolved at 7700, then resumed prediction. \\
    7719 & Missing 'fly16' & Resolved at 7719, then resumed prediction. \\
    7729 & Missing 'fly16' & Resolved at 7728, then resumed prediction. \\
    7793 & Missing 'fly32' & Resolved at 7793, then resumed prediction. \\
    8836 & Missing 'fly67' & Resolved at 8835, then resumed prediction. \\
    8848 & Missing 'fly77' & Resolved at 8847, then resumed prediction. \\
    8967 & Missing 'fly5' & Resolved at 8967, then resumed prediction. \\
    9054 & Missing 'fly34' & Resolved at 9053, then resumed prediction. \\
    9095 & Missing 'fly28' & Resolved at 9095, then resumed prediction. \\
    9161 & Missing 'fly12' & Resolved at 9160, then resumed prediction. \\
    9168 & Missing 'fly12' & Resolved at 9167, then resumed prediction. \\
    9240 & Missing 'fly4' & Resolved at 9239, then resumed prediction. \\
    9238 & Missing 'fly4' & Resolved at 9238, then resumed prediction. \\
    9258 & Missing 'fly77' & Resolved at 9258, then resumed prediction. \\
    9403 & Missing 'fly32' & Resolved at 9403, then resumed prediction. \\
    9607 & Missing 'fly32' & Resolved at 9606, then resumed prediction. \\
    9876 & Missing 'fly27' & Resolved at 9876, then resumed prediction. \\
    9891 & Missing 'fly21' & Resolved at 9889, then resumed prediction. \\
    10041 & Missing 'fly55' & Resolved at 9911, then resumed prediction. \\
    10099 & Missing 'fly34' & Resolved at 10097, then resumed prediction. \\
    10623 & Missing 'fly76' & Resolved at 10615, then resumed prediction. \\
    10710 & Missing 'fly79' & Resolved at 10710, then resumed prediction. \\
    10711 & Missing 'fly79' & Resolved at 10710, then resumed prediction. \\
    10917 & Missing 'fly2' & Resolved at 10911, then resumed prediction. \\
    10957 & Missing 'fly5' & Resolved at 10956, then resumed prediction. \\
    11434 & Missing 'fly79' & Resolved at 11433, then resumed prediction. \\
    12036 & Missing 'fly55' & Resolved at 12035, then resumed prediction. \\
    12259 & Missing 'fly61' & Resolved at 12250, then resumed prediction. \\
    12615 & Missing 'fly16' & Resolved at 12607, then resumed prediction. \\
    12615 & Missing 'fly16' & Resolved at 12615, then resumed prediction. \\
    12631 & Missing 'fly31' & Resolved at 12610, then resumed prediction. \\
    12615 & Missing 'fly16' & Resolved at 12615, then resumed prediction. \\
    12631 & Missing 'fly31' & Resolved at 12631, then resumed prediction. \\
    12671 & Missing 'fly32' & Resolved at 12670, then resumed prediction. \\
    12737 & Missing 'fly61' & Resolved at 12737, then resumed prediction. \\
    12753 & Missing 'fly61' & Resolved at 12751, then resumed prediction. \\
    12880 & Missing 'fly4' & Resolved at 12876, then resumed prediction. \\
    13123 & Missing 'fly40' & Resolved at 13123, then resumed prediction. \\
    13334 & Missing 'fly55' & Resolved at 13330, then resumed prediction. \\
    13369 & Missing 'fly62' & Resolved at 13368, then resumed prediction. \\
    13617 & Missing 'fly32' & Resolved at 13613, then resumed prediction. \\
    13752 & Missing 'fly8' & Resolved at 13713, then resumed prediction. \\
    13726 & Missing 'fly56' & Resolved at 13726, then resumed prediction. \\
    13750 & Missing 'fly8' & Resolved at 13749, then resumed prediction. \\
    13916 & Missing 'fly72' & Resolved at 13873, then resumed prediction. \\
    14022 & Missing 'fly8' & Resolved at 14021, then resumed prediction. \\
    14046 & Missing 'fly5' & Resolved at 14045, then resumed prediction. \\
    14146 & Missing 'fly5' & Resolved at 14144, then resumed prediction. \\
    14314 & Missing 'fly28' & Resolved at 14313, then resumed prediction. \\
    14482 & Missing 'fly71' & Resolved at 14482, then resumed prediction. \\
    14700 & Missing 'fly31' & Resolved at 14699, then resumed prediction. \\
    14737 & Missing 'fly3' & Resolved at 14735, then resumed prediction. \\
    14833 & Missing 'fly16' & Resolved at 14832, then resumed prediction. \\
    14862 & Missing 'fly55' & Resolved at 14845, then resumed prediction. \\
    14869 & Missing 'fly55' & Resolved at 14863, then resumed prediction. \\
    14867 & Missing 'fly55' & Resolved at 14866, then resumed prediction. \\
    14899 & Missing 'fly50' & Resolved at 14899, then resumed prediction. \\
    14948 & Missing 'fly0' & Resolved at 14945, then resumed prediction. \\
    15110 & Missing 'fly66' & Resolved at 15109, then resumed prediction. \\
    15229 & Missing 'fly8' & Resolved at 15229, then resumed prediction. \\
    15258 & Missing 'fly72' & Resolved at 15256, then resumed prediction. \\
    15591 & Missing 'fly8' & Resolved at 15589, then resumed prediction. \\
    15597 & Missing 'fly8' & Resolved at 15594, then resumed prediction. \\
    15759 & Missing 'fly10' & Resolved at 15757, then resumed prediction. \\
    15847 & Missing 'fly72' & Resolved at 15845, then resumed prediction. \\
    15951 & Missing 'fly4' & Resolved at 15950, then resumed prediction. \\
    16010 & Missing 'fly52' & Resolved at 15976, then resumed prediction. \\
    16010 & Missing 'fly52' & Resolved at 16009, then resumed prediction. \\
    16026 & Missing 'fly11' & Resolved at 16021, then resumed prediction. \\
    16074 & Missing 'fly10' & Resolved at 16071, then resumed prediction. \\
    16082 & Missing 'fly66' & Resolved at 16081, then resumed prediction. \\
    16102 & Missing 'fly66' & Resolved at 16100, then resumed prediction. \\
    16175 & Missing 'fly5' & Resolved at 16174, then resumed prediction. \\
    16202 & Missing 'fly63' & Resolved at 16200, then resumed prediction. \\
    16261 & Missing 'fly39' & Resolved at 16261, then resumed prediction. \\
    16279 & Missing 'fly32' & Resolved at 16266, then resumed prediction. \\
    16317 & Missing 'fly0' & Resolved at 16317, then resumed prediction. \\
    16354 & Missing 'fly4' & Resolved at 16347, then resumed prediction. \\
    16358 & Missing 'fly4' & Resolved at 16353, then resumed prediction. \\
    16445 & Missing 'fly67' & Resolved at 16445, then resumed prediction. \\
    16507 & Missing 'fly32' & Resolved at 16507, then resumed prediction. \\
    16655 & Missing 'fly12' & Resolved at 16654, then resumed prediction. \\
    16679 & Missing 'fly32' & Resolved at 16678, then resumed prediction. \\
    16680 & Missing 'fly28' & Resolved at 16674, then resumed prediction. \\
    16831 & Missing 'fly72' & Resolved at 16830, then resumed prediction. \\
    16910 & Missing 'fly66' & Resolved at 16907, then resumed prediction. \\
    17066 & Missing 'fly40' & Resolved at 17065, then resumed prediction. \\
    17158 & Missing 'fly5' & Resolved at 17158, then resumed prediction. \\
    17187 & Missing 'fly5' & Resolved at 17186, then resumed prediction. \\
    17277 & Missing 'fly66' & Resolved at 17273, then resumed prediction. \\
    17285 & Missing 'fly79' & Resolved at 17284, then resumed prediction. \\
    17338 & Missing 'fly50' & Resolved at 17318, then resumed prediction. \\
    17494 & Missing 'fly36' & Resolved at 17493, then resumed prediction. \\
    17601 & Missing 'fly65' & Resolved at 17600, then resumed prediction. \\
    17603 & Missing 'fly65' & Resolved at 17602, then resumed prediction. \\
    17649 & Missing 'fly0' & Resolved at 17645, then resumed prediction. \\
    17646 & Missing 'fly35' & Resolved at 17646, then resumed prediction. \\
    17649 & Missing 'fly0' & Resolved at 17649, then resumed prediction. \\
    17684 & Missing 'fly31' & Resolved at 17676, then resumed prediction. \\
    17691 & Missing 'fly4' & Fixed smaller polygon at 17690, then resumed prediction. \\  
    17861 & Missing 'fly7' & Fixed smaller polygon at 17834, then resumed prediction. \\  
    17882 & Missing 'fly79' & Fixed smaller polygon at 17880, then resumed prediction. \\ 
    17893 & Missing 'fly32' & Fixed smaller polygon at 17892, then resumed prediction. \\
    17933 & Missing 'fly16' & Fixed smaller polygon at 17932, then resumed prediction. \\
    17943 & Missing 'fly31' & Fixed smaller polygon at 17942, then resumed prediction. \\
    17955 & Missing 'fly31' & Fixed smaller polygon at 17953, then resumed prediction. \\
    18078 & Missing 'fly40' & Fixed smaller polygon at 18074, then resumed prediction. \\
    18110 & Missing 'fly4' & Fixed smaller polygon at 18076, then resumed prediction. \\
    18112 & Missing 'fly27' & Fixed smaller polygon at 18353, then resumed prediction. \\ 
    18112 & Missing 'fly27' & Fixed smaller polygon at 18353, then resumed prediction. \\
    18354 & Missing 'fly72' & Fixed smaller polygon at 18353, then resumed prediction. \\
    18475 & Missing 'fly9' & Fixed smaller polygon at 18473, then resumed prediction. \\
    18555 & Missing 'fly9' & Fixed smaller polygon at 18553, then resumed prediction. \\
    18687 & Missing 'fly4' & Resolved at 18686, then resumed prediction. \\
    18735 & Missing 'fly72' & Resolved at 18689, then resumed prediction. \\
    18738 & Missing 'fly72' & Resolved at 18730, then resumed prediction. \\
    18830 & Missing 'fly32' & Resolved at 18826, then resumed prediction. \\
    18876 & Missing 'fly9' & Resolved at 18873, then resumed prediction. \\
    18929 & Missing 'fly9' & Resolved at 18928, then resumed prediction. \\
    18959 & Missing 'fly12' & Resolved at 18959, then resumed prediction. \\
    19025 & Missing 'fly71' & Resolved at 19021, then resumed prediction. \\
    19071 & Missing 'fly66' & Resolved at 19070, then resumed prediction. \\
    19084 & Missing 'fly4' & Resolved at 19083, then resumed prediction. \\
    19148 & Missing 'fly39' & Resolved at 19146, then resumed prediction. \\
    19201 & Missing 'fly39' & Resolved at 19179, then resumed prediction. \\
    19204 & Missing 'fly39' & Resolved at 19202, then resumed prediction. \\
    19376 & Missing 'fly8' & Resolved at 19373, then resumed prediction. \\
    19441 & Missing 'fly9' & Resolved at 19441, then resumed prediction. \\
    19444 & Fixed a enlarged polygon & Resolved at 19444 , then resumed prediction. \\
    19513 & Missing 'fly16' & Resolved at 19513, then resumed prediction. \\
    19523 & Fixed a enlarged polygon & Resolved at 19523, then resumed prediction. \\
    19732 & Missing 'fly9' & Resolved at 19731, then resumed prediction. \\
    19961 & Missing 'fly39' & Resolved at 19959, then resumed prediction. \\
    20039 & Missing 'fly66' & Resolved at 20039, then resumed prediction. \\
    20110 & Missing 'fly66' & Resolved at 20109, then resumed prediction. \\
    20141 & Missing 'fly40' & Resolved at 20139, then resumed prediction. \\
    20269 & Missing 'fly26' & Resolved at 20269, then resumed prediction. \\
    20289 & Missing 'fly32' & Resolved at 20288, then resumed prediction. \\
    20338 & Missing 'fly48' & Resolved at 20336, then resumed prediction. \\
    20370 & Missing 'fly8' & Resolved at 20352, then resumed prediction. \\
    20371 & Missing 'fly8' & Resolved at 20367, then resumed prediction. \\
    20422 & Missing 'fly10' & Resolved at 20420, then resumed prediction. \\
    20493 & No issue  & Video completed \\
    \hline
\end{longtable}



\end{document}